\documentclass{IEEEtaes}

\usepackage[colorlinks,urlcolor=blue,linkcolor=blue,citecolor=blue]{hyperref}
\usepackage{bookmark}
\usepackage{color,array,amsthm}
\usepackage{cite}
\usepackage{graphicx}
\usepackage{upgreek}
\usepackage{booktabs}  
\usepackage{threeparttable}  
\usepackage{multirow}
\usepackage{epstopdf}
\usepackage{amssymb}
\usepackage{algorithm}
\usepackage{algorithmicx}
\usepackage{algpseudocode}
\usepackage{setspace}
\usepackage[cmex10]{amsmath}
\usepackage{url}
\usepackage{subfig}
\usepackage{gensymb}
\usepackage{amsthm}

\jvol{XX}
\jnum{XX}
\jmonth{XXXXX}
\paper{1234567}
\pubyear{2020}
\doiinfo{TAES.2020.Doi Number}

\newtheorem{lemma}{Lemma}
\begin{document}

\title{Space Non-cooperative Object Active Tracking with Deep Reinforcement Learning}

\author{Dong Zhou}

\author{Guanghui Sun}
\member{Senior Member, IEEE}
\author{Wenxiao Lei}
\affil{Harbin Institute of Technology, Harbin, China} 

\receiveddate{This work was kindly supported by the National Key R\&D Program of China through grant 2019YFB1312001.}

\corresp{}

\authoraddress{D. Zhou, G. Sun and W. Lei are with the Department of Control Science and Engineering, Harbin Institute of Technology, Harbin, China, 150001. E-mail:  dongzhou@hit.edu.cn, guanghuisun@hit.edu.cn. \itshape(Corresponding author: Guanghui Sun) }

\markboth{ZHOU ET AL.}{SPACE NON-COOPERATIVE OBJECT ACTIVE TRACKING}
\maketitle

\begin{abstract}
	Active visual tracking of space non-cooperative object is significant for future intelligent spacecraft to realise space debris removal,  asteroid exploration, autonomous rendezvous and docking. However, existing works often consider this task into different subproblems (e.g. image preprocessing, feature extraction and matching, position and pose estimation, control law design) and optimize each module alone, which are trivial and sub-optimal. To this end, we propose an end-to-end active visual tracking method based on DQN algorithm, named as DRLAVT. It can guide the chasing spacecraft approach to arbitrary space non-cooperative target merely relied on color or RGBD images, which significantly outperforms position-based visual servoing baseline algorithm that adopts state-of-the-art 2D monocular tracker, SiamRPN. Extensive experiments implemented with diverse network architectures, different perturbations and multiple targets demonstrate the advancement and robustness of DRLAVT. In addition, We further prove our method indeed learnt the motion patterns of target with deep reinforcement learning through hundreds of trial-and-errors.
\end{abstract}


\section{INTRODUCTION}
S{\scshape pace} non-cooperative object visual tracking, as one of the key components of vision system on spacecraft, has wide applications in space debris removal, malfunctioning spacecraft maintenance, asteroid exploration, and autonomous rendezvous and docking \cite{huangDexterousTetheredSpace2017, flores-abadReviewSpaceRobotics2014, dongAutonomousRoboticCapture2016,petitVisionbasedDetectionTracking2012, ramosVisionbasedTrackingNoncooperative2018,zhou3DVisualTracking2021}. 

According to the presence of control command, space non-cooperative object visual tracking can be concluded into passive and active visual tracking. Many of efforts have been devoted to studying passive methods \cite{ fourieFlightResultsVisionBased2014, zhangImprovedRealtimeVisual2016, volpePassiveCameraBased2018, zhou2DVisionbasedTracking2021}. Fourie \cite{fourieFlightResultsVisionBased2014} put forward visual inspection and tracking algorithm, VERTIGO, by using stereo camera, of which effectiveness had been proven through on-orbit experiment implemented in International Space Station. 
To remove geostationary orbit space debris, a monocular real-time robust feature tracking algorithm was proposed in the paper \cite{huangDexterousTetheredSpace2017} that combines SURF keypoints detector and Pyramid-Kanade-Lucas-Tomasi (P-KLT) matching method. In our preliminary work \cite{zhou2DVisionbasedTracking2021}, we introduced plenty of general visual tracking algorithms rooted in computer vision society to aerospace domain, which demonstrate their powerful performances to track space non-cooperative objects with 2D image. However, all the passive visual tracking methods are easily failed, because of complex 6-Degrees-of-Freedom (DoF) motion in large range of space non-cooperative objects and low-resolution vision sensors with small FOV equipmented on spacecraft. It dramatically shortens observation period and severely blocks subsequent missions to proceed.

\begin{figure}
	\centering
	\includegraphics[width=0.45 \textwidth]{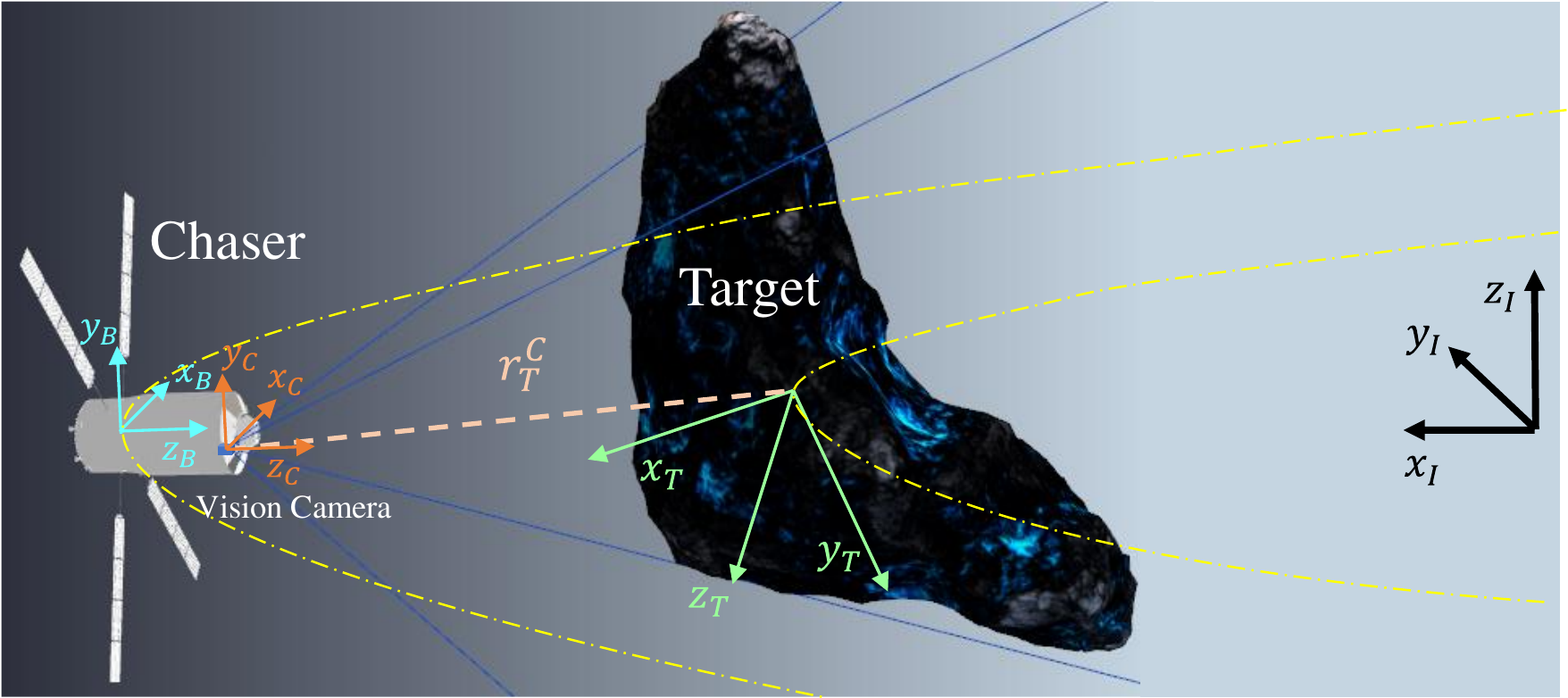}
	\centering
	\caption{The schematic of space non-cooperative object active visual tracking. The yellow dash lines denote the orbits of chaser spacecraft and target respectively. The pyramid composed by blue solid line represents the FOV of vision camera equipmented on chaser.}
	\label{fig1}
\end{figure}

\begin{figure*}
	\centering
	\includegraphics[width=0.7 \textwidth]{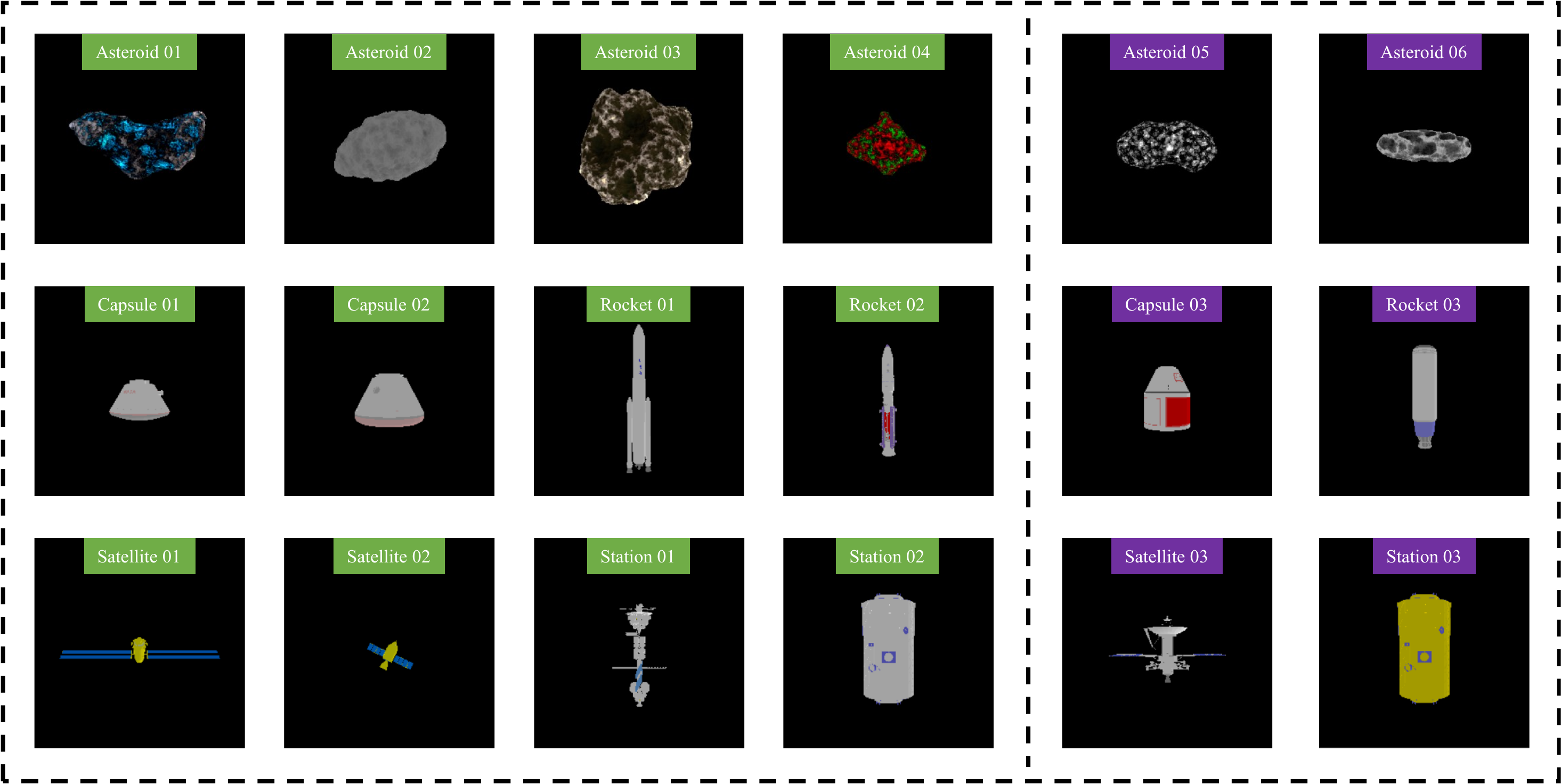}
	\caption{The 18 types of space non-cooperative target models, involving Asteroids, Return Capsules, Rockets, Satellites and Space Station. Objects with green labels are utilized for training DRLAVT model, the others are adopted to evaluate active visual trackers.}
	\label{fig2}
\end{figure*}

To this end, space non-cooperative object active visual tracking has garnered much more concerns, also considered as visual servoing, which stays object in the field of view or makes the relative position and attitude between chaser (e.g. spacecraft and robotic manipulator) and target to an expected state with vision sensors. It can provide more relaxed condition for follow-up tasks such as space debris capture and removal, autonomous rendezvous and docking, and orbiting observation and exploration. In general, active visual tracking is typically classified into two main categories: Position-based Visual Servoing (PBVS) \cite{dongPositionbasedVisualServo2015a, dongAutonomousRoboticCapture2016, sunAdaptiveRelativePose2018, liuRobustAdaptiveRelative2020} and Image-based Visual Servoing (IBVS) \cite{wangEyeinHandTrackingControl2017, felicettiImagebasedAttitudeManeuvers2018 ,zhaoImagebasedControlRendezvous2021}. 

The essence of PBVS methods is that visual perception and controller design are solved respectively. It can directly and naturally navigate chaser to the target with expected relative pose. Dong \cite{dongAutonomousRoboticCapture2016} proposed a PBVS scheme for space robotic manipulator to capture non-cooperative object, in which photogrammetry and Adaptive Extended Kalman Filter (AEKF) were combined to estimate 6-DoF pose of target. However, this method was merely evaluated under an easy simulated configuration where the non-cooperative object possesses specific and discriminative image features with simple motion patterns. Sun \cite{sunAdaptiveRelativePose2018} and Liu \cite{liuRobustAdaptiveRelative2020} respectively presented robust adaptive PBVS controllers with sliding mode theory, assuming that the pose of non-cooperative target is accurately measured during approaching stage. In real application, it is absolutely hard to achieve this ideal precondition, which makes most of PBVS algorithms weak and vulnerable. We therefore introduce a simple but powerful PBVS method as baseline algorithm that adopts state-of-the-art 2D monocular tracker (e.g. KCF\cite{henriquesHighSpeedTrackingKernelized2015}, SiamFC\cite{liHighPerformanceVisual2018}) with RGB-D image to predict 3D position of any non-cooperative target. 

All the methods mentioned before, whatever PBVS or IBVS algorithms, decompose active visual tracking task into many sub-problems including features extraction, features matching, pose estimation, control law design. Many researchers \cite{levineLearningContactrichManipulation2015, levineEndtoendTrainingDeep2016, loquercioDeepDroneRacing2020} have proposed that method following the idea of task decomposition is often trivial and suboptimal while it can not adapt to complex and dynamic environment. With the rapid development of deep learning, deep reinforcement learning that train an end-to-end neural network with large-scale experimental data sampled by trial-and-error can learn global optimal policy \cite{suttonReinforcementLearningIntroduction2018}. This provide a novel perspective to solve active visual tracking task. 

In recent, deep reinforcement learning has made plenty of achievements in aerospace domain. Gaudet \cite{briangaudetAdaptivePinpointFuel2014} utilized reinforcement learning to realise fuel-efficient Mars landing solution. Scorsoglio \cite{scorsoglioImagebasedDeepReinforcement} also presented an adaptive landing method based on Proximal Policy Optimization (PPO) for lunar pinpoint landing by using image and altimeter measurement as input. Sanchez \cite{sanchez-sanchezRealTimeOptimalControl2018} trained deep neural networks to approximate optimal solution, which was validated on the pinpoint landing problem of mass varing spacecraft with bounded thrust. High-level mission planning and decision-making of spacecraft based on the combination of Partially-Observable Markov Decision Process (POMDP) with Deep Q-learning \cite{mnihHumanlevelControlDeep2015} were presented in paper \cite{harris2019spacecraft}. Hovell \cite{hovellDeepReinforcementLearning2020, kirkhovellDeepReinforcementLearning2021} adopted D4PG algorithm \cite{barth-maronDistributedDistributionalDeterministic2018} as deep guidance of conventional controller for spacecraft approaching and docking, which alleviated the simulation-to-reality gap problem \cite{bousmalisUsingSimulationDomain2018,sunderhaufLimitsPotentialsDeep2018} and achieved comparable performance on planar gravity-offset testbed. However, most of related works in aerospace domain have not considered learning policy directly from image data of which information capacity is much larger than unreliable states of space non-cooperative target computed by trivial estimation algorithm (e.g. object position, velocity, attitude, and angular velocity). 

Therefore, inspired by the method in \cite{luoEndtoEndActiveObject2020} that can only track simple translational target with discriminative features in normal 2D planar environment, we propose an end-to-end space non-cooperative object active visual tracking algorithm, named as DRLAVT, which learns 3D visual servoing policy straightforward from extensive experimental images sampled by trial-and-error with our simulation platform. Comparing to the PBVS baseline, DRLAVT achieves excellent tracking performance even under severe perturbations like actuator noise, time-delay and image blur.

The contributions of our work in this paper are summarised as following: (1) An end-to-end active visual tracker based on deep Q-learing algorithm, named as DRLAVT, is presented. It provides interesting and pioneering methodology for space non-cooperative object tracking, which learns optimal policy from raw color image or RGBD image directly. (2) As comparison, we propose a novel and robust PBVS baseline method that adopts state-of-the-art 2D monocular tracking algorithm with RGBD images to predict 3D position of target. (3) To train and evaluate active visual trackers, we construct simulation environment that consists of 18 space non-cooperative object models and a chaser spacecraft equipmented with vision sensors, which is available and open-source on \url{https://github.com/Dongzhou-1996/SNCOAT}.

The remaining paper is organized as follows: Section \ref{section2} formulates the problem of active visual tracking for space non-cooperative object and introduces the configurations of simulation environment. We propose PBVS baseline algorithm in Section \ref{section3} and our DRLAVT algorithm in Section \ref{section4}. Furthermore, extensive experiments are implemented in Section \ref{section5} to show the effectiveness and advancement of DRLAVT. At final, Section \ref{section6} concludes this paper.

\section{Problem Formulation and Simulation Environment \label{section2}}
\subsection{Active Visual Tracking Problem \label{section2_1}}
An active visual tracking problem involves a chaser spacecraft equipmented with vision camera and a non-cooperative object (e.g. spacecrafts, asteroids, rockets) with complex 6-DoF motion, which is shown in Fig \ref{fig1}. It is worthy noting that we only focus on how to guide spacecraft approach to the target with image, that is, no relative attitude synchronization is considered in this paper. Besides, we further simplify the dynamic models of chaser and target \cite{hovellDeepReinforcementLearning2020,kirkhovellDeepReinforcementLearning2021} by an assumption that both objects are free-floating under close range (less than 50m), although more precise orbital dynamic models are adopted in other studies. 

During active visual tracking, Four coordinate systems depicted in Fig \ref{fig1} are involved: 
\begin{itemize}
	\item Reference frame, denoted by $\mathcal{F}_{I}$, of which origin is located at somewhere near to both chaser and target;
	\item Chaser body-fixed frame, denoted by $\mathcal{F}_{B}$, of which origin is the Center of Mass (CoM) of spacecraft;
	\item Vision camera frame, denoted by $\mathcal{F}_{C}$, which is axis-aligned with $\mathcal{F}_{B}$. The transformation matrix from $\mathcal{F}_{C}$ to $\mathcal{F}_{B}$ is denoted as $M^{B}_C$;
	\item Target body-fixed frame, denoted by $\mathcal{F}_{T}$, of which origin is the CoM of target;
\end{itemize}

We first give the simplified dynamic model of chasing spacecraft:
\begin{equation}
	\ddot{X}^{I}_{s} = \frac{u}{m_s} + \Omega
\end{equation}
in which, ${X}^{I}_{s}$ is the 3D position of chaser (i.e. CoM of chaser) in $\mathcal{F}_{I}$. $u = \{F_x, F_y, F_z\}$ is the control force, In real application, the control force provided by thruster is often bounded. To this end, we assume that $\left\lVert u \right\rVert_{\infty } \leq 50$N. $\Omega$ is noise term introduced by actuator distrubation, computational time-delay and image blur together. In addition, $m_s$ is the mass of spacecraft. It is set to 113.9kg.

And then, because of the non-cooperative nature of target, we consider there are no control force and torque applied on it. Hence that, complex 6-DoF motion is reduced to uniform linear motion and rotation:
\begin{align}
	\dot{X}^{I}_{T} = C_{pos} \\
	\nonumber \dot{\omega}^{I}_T = C_{ang}
\end{align}
where ${X}^{I}_{T} = \{x^{I}_T, y^{I}_T, z^{I}_T\}$ and ${\omega}^{I}_T = \{ \alpha, \beta, \gamma \}$ are respectively the 3D position and angular of target in ${\mathcal{F}_I}$. $C_{pos}$ and $C_{ang}$ are two constants denoted the translation and rotation speed. 

The goal of active visual tracking can be formulated as follows:
\begin{equation}
	e = \left\lVert r^{B}_{T} - r^{\ast} \right\rVert_2 =  \left\lVert M^{B}_{C} \cdot r^{C}_{T} - r^{\ast} \right\rVert_2 = 0 \label{eq_3}
\end{equation}
in which, $r^{B}_{T}$ is the position of target in $\mathcal{F}_B$, $r^{\ast}$ denotes the expected distance between chaser and target. In this work, we set $r^{\ast} = \{0, 0, 5\}$.

\begin{figure}
	\centering
	\subfloat[Color image]{\includegraphics[width=0.2\textwidth]{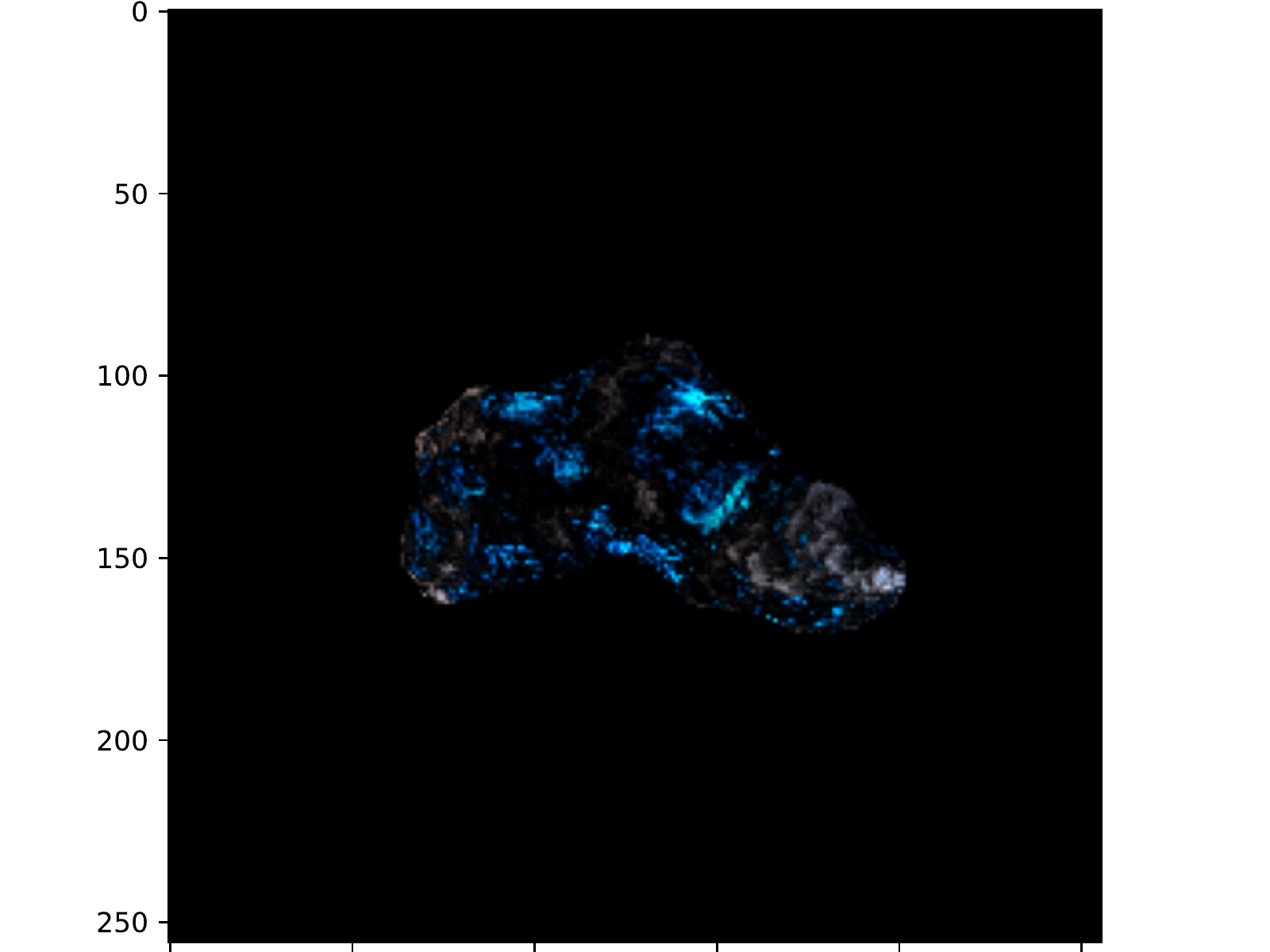}} \hfil
	\subfloat[Depth map]{\includegraphics[width=0.2\textwidth]{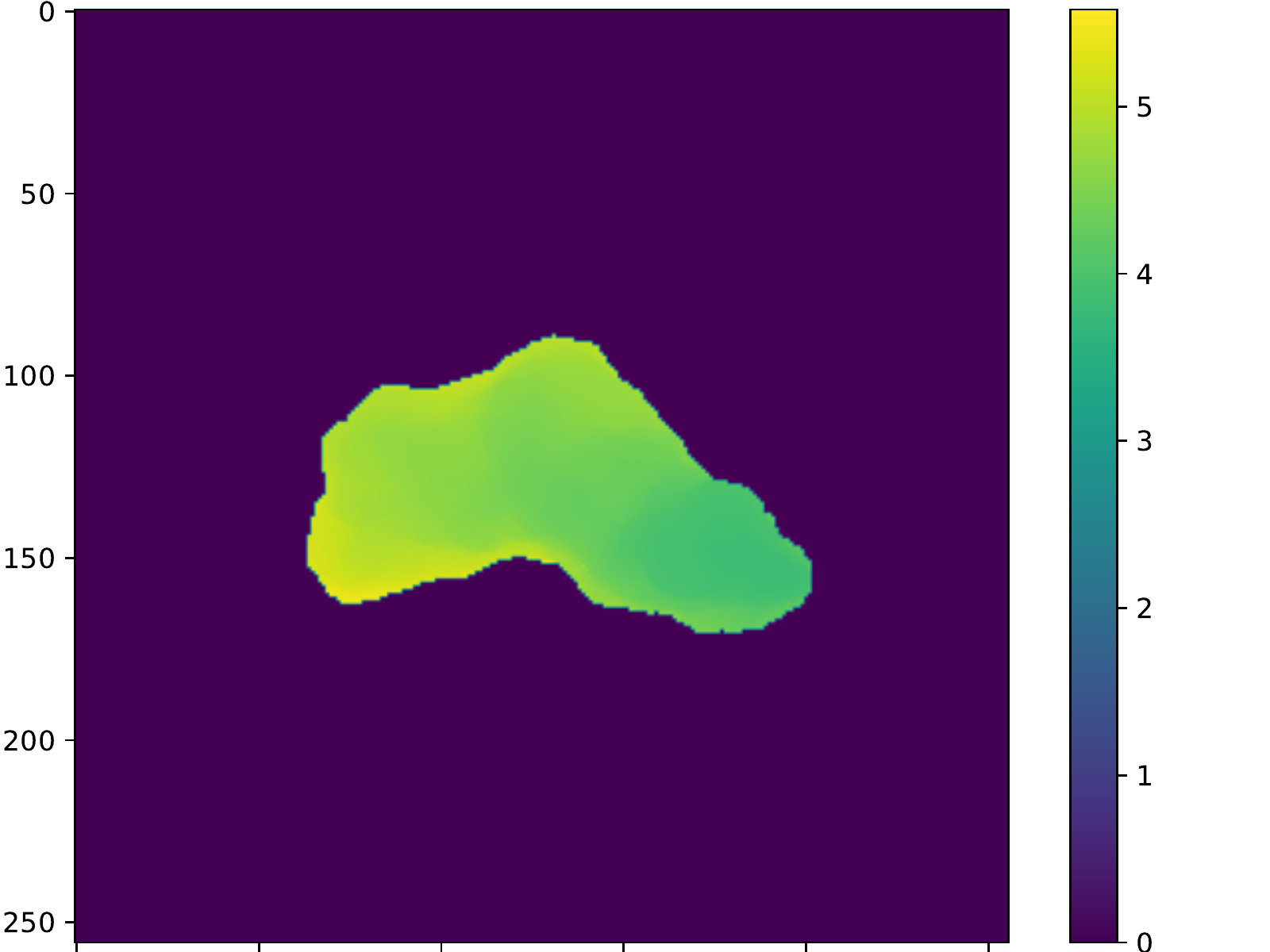}}
	\caption{Color image and depth map of Asteroid 01 simultaneously captured by simulated vision camera.}
	\label{fig3}
\end{figure}

\begin{figure}
	\centering
	\includegraphics[width=0.3\textwidth]{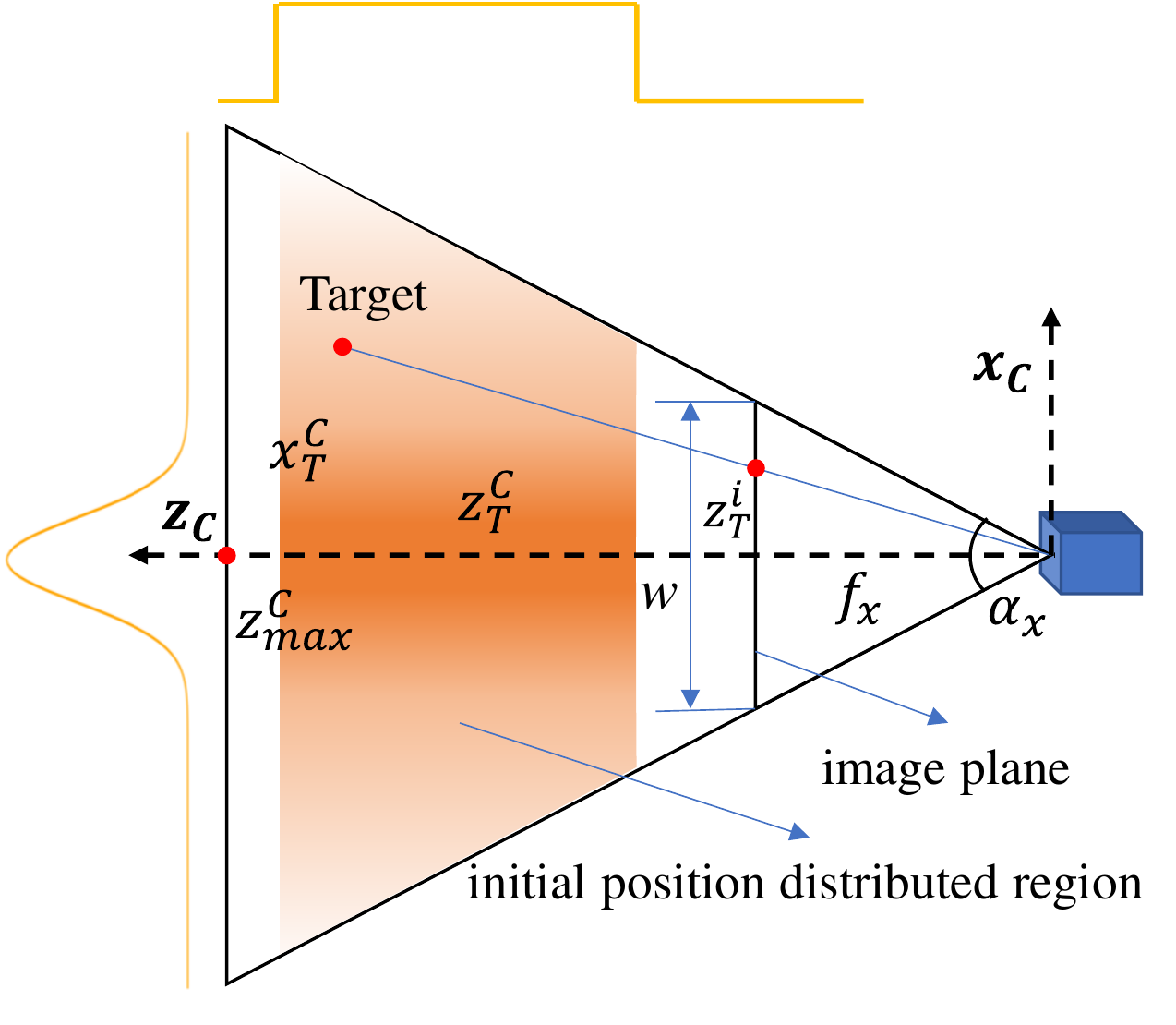}
	\caption{The schematic of perspective vision camera in simulated environment. We also depict the distribution of initial position of non-cooperative target in $\mathcal{F}_C$.}
	\label{fig4}
\end{figure}

\subsection{Environment Specification \label{section2_2}}
To study more powerful and general active visual tracking algorithm, especially for deep reinforcement learning based methods, we construct a space simulation environment by virtual physics engine, CoppeliaSim, which includes spacecraft equipmented with a perspective vision camera and 18 types of non-cooperative object models. All the models can be concluded into 5 categories: (1) Asteroids, (2) Return Capsules, (3) Rockets, (4) Satellites, (5) Space stations. It is clearly shown in Fig \ref{fig2} that the geometry, size, and texture of different types of targets are various. Therefore, traditional PBVS and IBVS algorithms based on hand-crafted image features are impossible to be adapted to all the objects. 

We suppose that the only visual sensor in simulated environment to perceive non-cooperative target is a perspective vision camera installed at the front of chaser as shown in Fig \ref{fig1}. However, it is worthwhile noting that both color image and depth map (see in Fig \ref{fig3}) can be achieved by it. The model of vision camera is described in Fig \ref{fig4}, which is important for active visual tracking. Due to only perspective angle $(\alpha_x, \alpha_y)$ and resolution $\left(W, H\right)$ are known in CoppeliaSim, we therefore compute camera intrinsic matrix $M_{intr}$ following Eq. \ref{eq_4}:
\begin{equation}
    M_{intr}=
    \begin{bmatrix}
       \frac{W}{2 \tan(\alpha_x/2)} & 0 & \frac{W}{2} \\
       0 &   \frac{H}{ 2 \tan(\alpha_y/2)} & \frac{H}{2}\\
       0 & 0 & 1
    \end{bmatrix} 
    \label{eq_4}
 \end{equation}
 In this work, we set $\alpha_x = \alpha_y = 60 \degree$ and $W = H = 256$. 

Furthermore, the initial state of non-cooperative object is another important configuration for active visual tracking research, involving initial position, attitude, velocity, and angular velocity. There is a necessary condition that target should be observable at beginning stage of tracking. To this end, we randomize the initial position ${X}^{C}_{T}(0) = \{x^{C}_{T}(0), y^{C}_{T}(0), z^{C}_{T}(0)\}$ of target with different distributions. Both $x^{C}_{T}(0)$ and $y^{C}_{T}(0)$ obey normal distribution $\mathcal{N}(0, 1)$, while $z^{C}_{T}(0)$ follows uniform distribution $U(2, 12)$. Referring to real situations and the configuration in other works, the initial velocity of non-cooperative target is randomized by $\mathcal{N}(0, 0.3)$. Because of the neglect of pose control as we assumed before, we simply assign the initial attitude $\omega^{I}_{T}$ and angular velocity $C_{ang}$ with random but reasonable value. 

\begin{figure}
	\centering
	\includegraphics[width=0.3\textwidth]{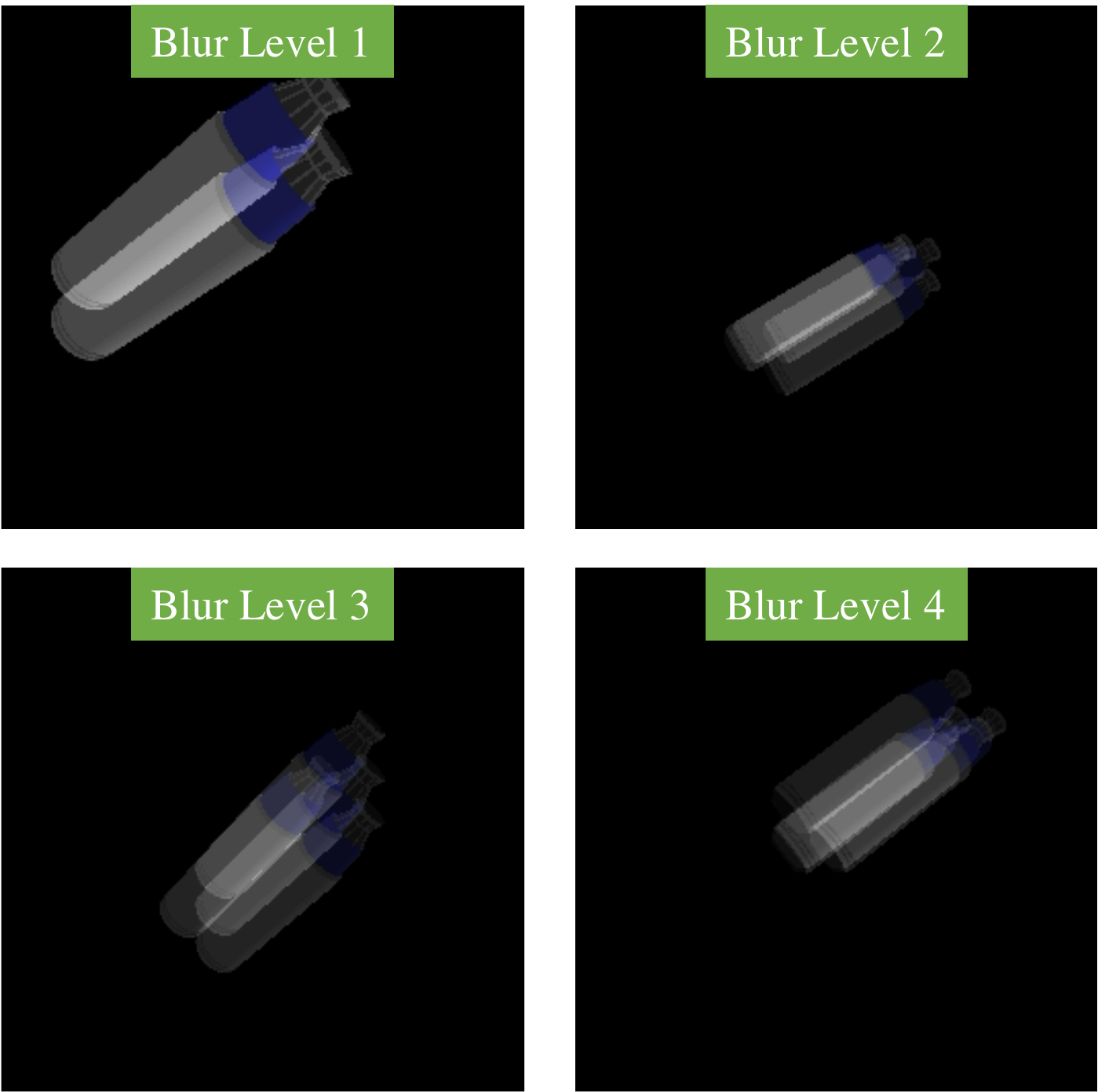}
	\caption{Four different levels of synthetic image blur.}
	\label{fig5}
	\centering
\end{figure}

\begin{figure*}
	\centering
	\includegraphics[width=0.6\textwidth]{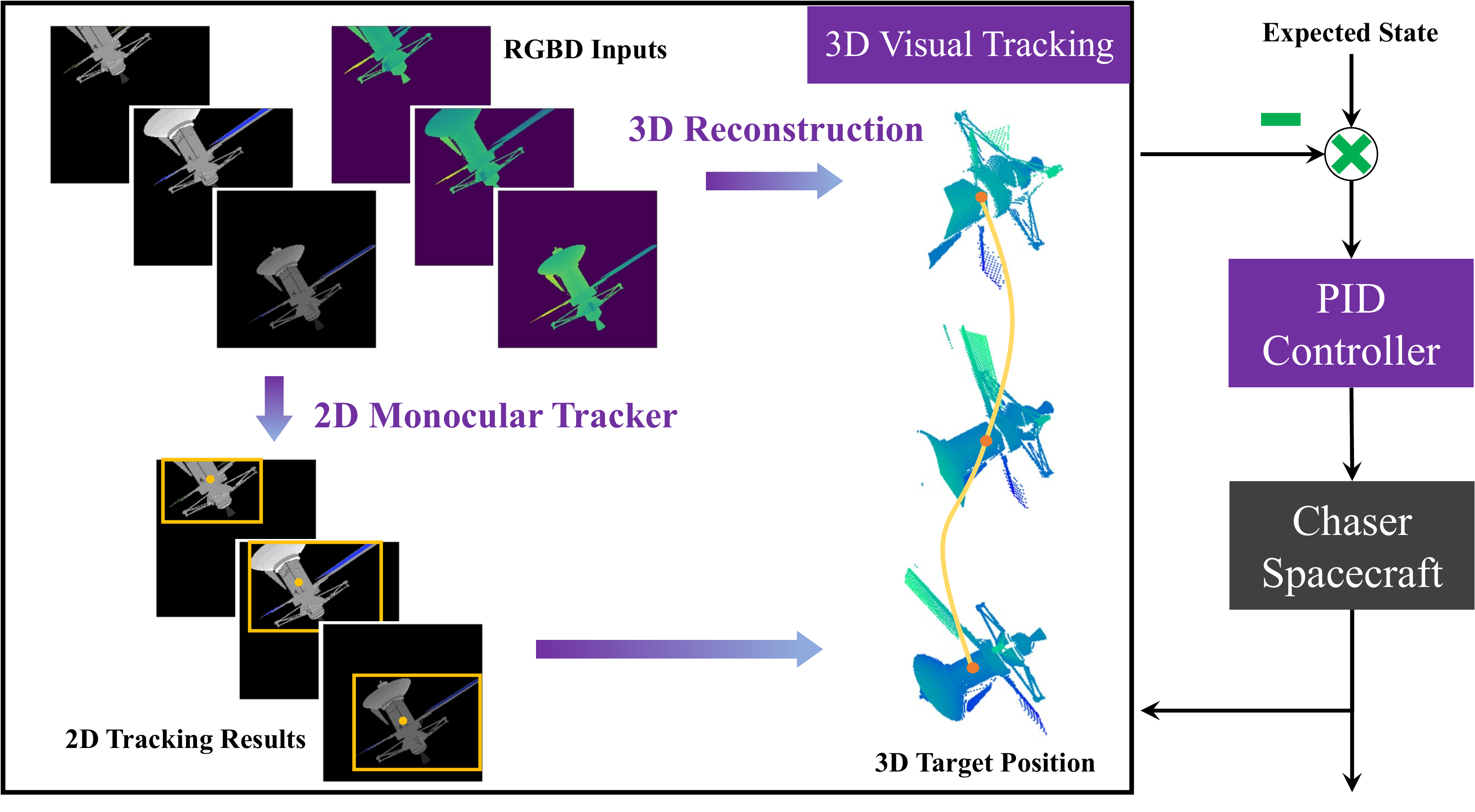}
	\caption{The framework of PBVS baseline algorithm}
	\label{fig6}
\end{figure*}

To validate the robustness of active visual trackers, three types of perturbations are introduced into our simulated environment. 
\begin{itemize}
	\item Actuator noise. In application, the margin between ideal controller output and real actuator output is inevitable. Therefore, we multiply action command with a noise factor $f_a$ which obeys $\mathcal{N}(1, 0.3)$.
	\item Processing time-delay. If active visual tracking algorithm is deployed on actual platform, the variance of running speeds will be another distinct disturbance. So we add a random time-delay $D$ to sampling time $\Delta t$ of simulation (original sampling time is set to 0.1s).
	\item Image blur. Because of rapid camera motion during active tracking, severe image blur often leads to tracking failed. In this paper, we simulate different levels of image blur by the average of N consecutive images ($N \geq 2$), formulated as:
	\begin{align}
		\tilde{I}^{N}_t &= \frac{1}{N} \sum_{i=0}^{N-1} {I_{t-i}}
	\end{align}
	in which, $I_t$ and $\tilde{I}^{N}_t$ are separately clear and blurred image acquired at t-th timesteps from simulated environment.
\end{itemize}

\subsection{Evaluation mechanism \label{section2_3}}
Reasonable evaluation mechanism is important to show real ability of active visual trackers. In this subsection, more details of our evaluation configuration are given. 

At first, one-shot protocol \cite{huangGOT10kLargeHighDiversity2019} is introduced into our evaluation, that is, the use of unseen classes of space non-cooperative objects to evaluation. As we mentioned before, there are 18 different types of target models constructed by us (see in Fig \ref{fig2}), in which two-thirds of model are used for training and the others are utilized for validation.

Second, we add delayed ending stage into each episode to explore recovery ability of active visual trackers, which can also balance the distribution of positive and negative samples during training. The length of delayed ending is recommended to be 10 $\sim$ 20 frames. 

Finally, the Average Episode Length (AEL) and Average Episode Reward (AER) are adopted as evaluation metrics. The higher AEL and AER are, the more accurate and robust active visual tracker is. AEL and AER are defined as follows:
\begin{align}
	AEL = \frac{1}{N\times R}\sum^{N}_n \sum^{R}_r L^{n}_{r} \\
	AER = \frac{1}{N\times R}\sum^{N}_n \sum^{R}_r R^{n}_{r}
\end{align}
where $L^{n}_{r}$ and $R^{n}_{r}$ denote episode length and reward of n-th object category at r-th evaluation. $N$ and $R$ are the number of classes and repetition times of evaluation, respectively. In this paper, $N = 5$ and $R = 20$. 

\begin{figure}
	\centering
	\includegraphics[width=0.48\textwidth]{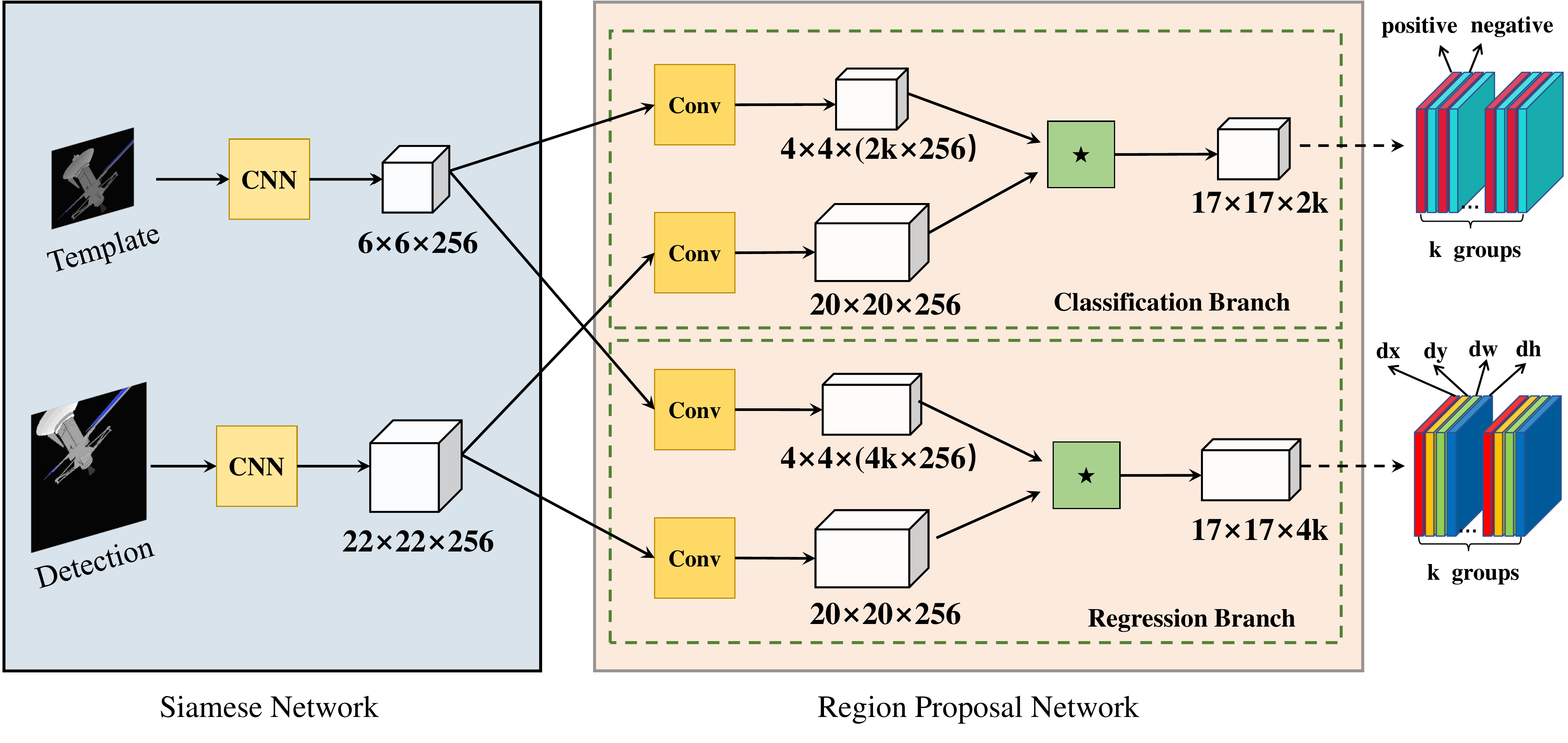}
	\caption{The framework of SiamRPN.}
	\label{fig7}
\end{figure}

\section{PBVS Baseline Algorithm \label{section3}}
In this section, we propose a novel PBVS baseline algorithm as comparison in later experiments, which mainly consists of 3D visual tracking module and conventional PID controller. The framework of our method is depicted in Fig \ref{fig6}. More details of this algorithm are described as follows. 

\begin{figure}
	\centering
	\includegraphics[width=0.25\textwidth]{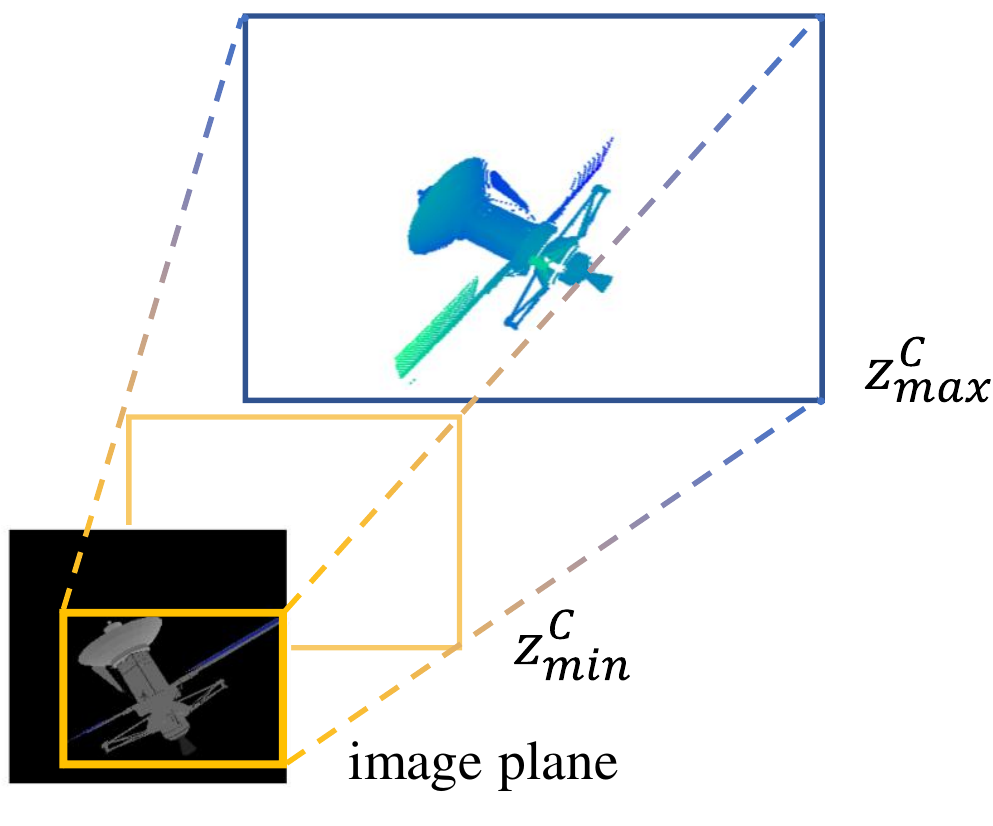}
	\caption{The schematic of frustum proposal extraction}
	\label{fig8}
\end{figure}

\begin{figure*}[ht]
	\centering
	\subfloat[ConvNet]{\includegraphics[width=0.6\textwidth]{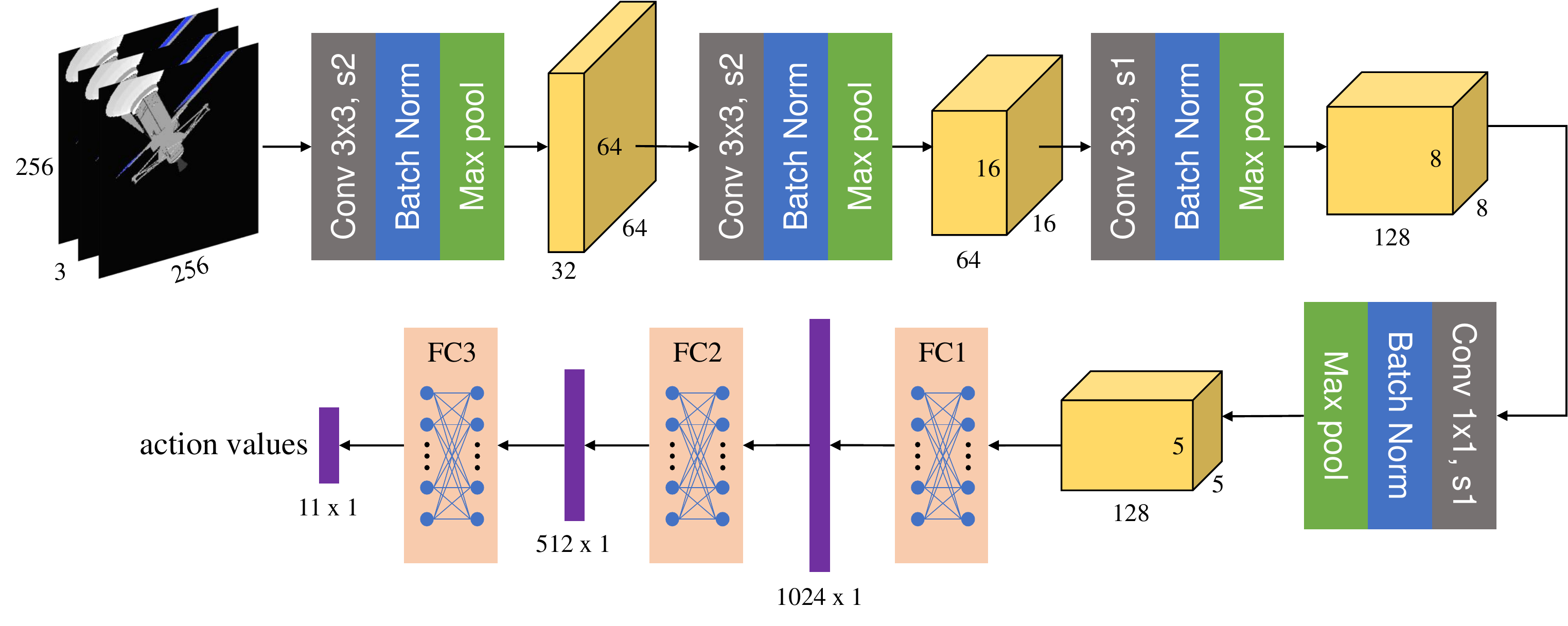} \label{fig9_1}} \vfil
	\subfloat[ResNet]{\includegraphics[width=0.6\textwidth]{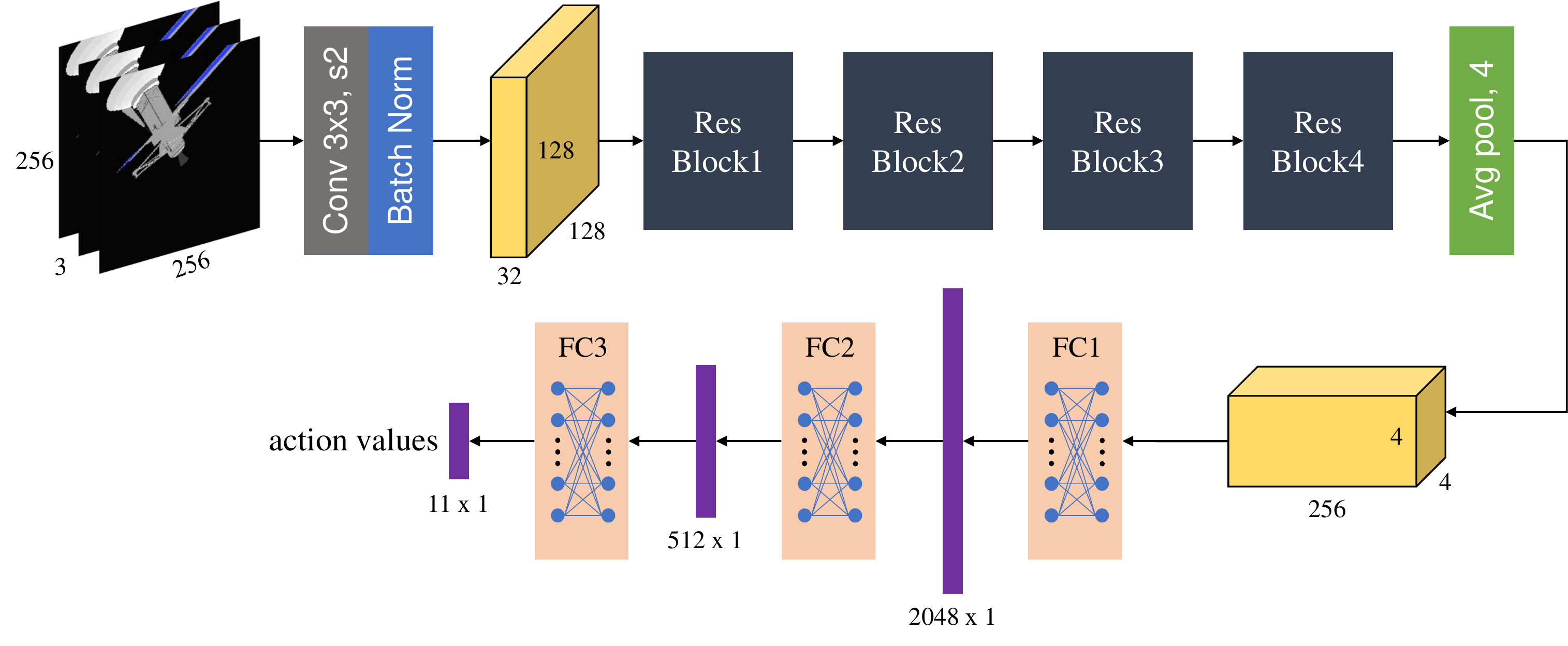} \label{fig9_2}} 
	\caption{Two types of deep neural network architectures adopted as Q-network in this work.}
	\label{fig9}
\end{figure*}

\subsection{3D Visual Tracking \label{section3_1}}
3D visual tracking is to perceive the 3D position of non-cooperative target, given the initial state of object. In this paper, the initial state is the 2D bounding-box of object in image space at the beginning of tracking, which is used for 2D monocular tracker. The process of 3D visual tracking is clearly shown in the left of Fig \ref{fig6}. We firstly proceed 2D monocular tracking and 3D reconstruction in parallel. Once the 2D tracking result and 3D point cloud of target are acquired, we further estimate the 3D target position by frustum average operation. It is worthwhile noting that 2D monocular tracking only takes color image as input. 

The state-of-the-art 2D monocular tracker, SiamRPN \cite{liHighPerformanceVisual2018}, is adopted into our framework, which combines deep Siamese network and Region Proposal Network (RPN) to accurately predict the 2D bounding-box of target, denoted as $\mathcal{A}_t = (x, y, w, h)$,  in detection image by using template image patch. The performance of SiamRPN in normal situations have been proven on different visual object tracking benchmarks, such as OTB \cite{wuObjectTrackingBenchmark2015}, VOT \cite{kristanSeventhVisualObject2019}, and GOT10K \cite{huangGOT10kLargeHighDiversity2019}. We also validate this tracker with SNCOVT dataset in our preliminary work, which demonstrated its generalization ability in aerospace domain \cite{zhou2DVisionbasedTracking2021}.

At the same time, we generate 3D point cloud $P_t =\{(x_i^C, y_i^C, z_i^C) \in \mathbb{R}^3 | i = 1, 2, ..., r\}$ with RGBD image, which can be easily calculated by following equation:
\begin{align}
	{z^C_i}\left[ {\begin{array}{*{20}{c}}
		u_i\\
		v_i\\
		1
		\end{array}} \right] = {M_{intr}}\left[ {\begin{array}{*{20}{c}}
		{{x^C_i}}\\
		{{y^C_i}}\\
		{{z^C_i}}
		\end{array}} \right]
\end{align}
in which, $M_{intr}$ is the intrinsic matrix of vision camera mentioned in Eq \ref{eq_4} and  ${z^C_i} > 0$. 

After 2D tracking result $\mathcal{A}_t$ and point cloud $P_t$ are both available, the frustum proposal $P_{frustum} =\{(x_i^C, y_i^C, z_i^C) \in \mathbb{R}^3 | i = 1, 2, ..., n\}$ can be extracted as shown in Fig \ref{fig8}. In this work, we assume the extracted frustum proposal only contains the point cloud of object of interest. So the 3D target position $r^{C}_{T}$ in $\mathcal{F}_C$ is estimated by using frustum average operation which is formulated as:
\begin{align}
	r_T^C   = \left[ {\begin{array}{*{20}{c}}
		{\frac{1}{n}\sum\limits_i^n {x_i^C} }, \quad
		{\frac{1}{n}\sum\limits_i^n {y_i^C} },	\quad
		{\frac{1}{n}\sum\limits_i^n {n_i^C} }
		\end{array}} \right]^{T}
\end{align}

\subsection{Controller Design \label{section3_2}}
Although various control methods (e.g. optimal control, sliding-mode control, robust control) for active visual tracking have been proposed, in our opinion, the accurate and real-time vision perception algorithm described hereinbefore will significantly alleviate complexity of control laws design. We therefore utilize classical PID controller to realise active visual tracking:
\begin{align}
	u(t) = K_{P} \cdot e(t) + K_{I} \cdot \int e(t)\,dt+ K_{D} \cdot \dot{e}(t)
\end{align}
where $e(t) = \left\lVert r^B_{T}(t) - r^{\ast}\right\rVert$, $K_{P}$, $K_{I}$ and $K_{D}$ are the coefficients of PID controller respectively. To correctly evaluate PBVS algorithm, we make a deal that the coefficients of controller are prohibited to manually fine-tune.

\section{DRLAVT Algorithm \label{section4}}
In this section, we propose an active visual tracker based on DQN, named as DRLAVT, which guides chaser spacecraft approach to the non-cooperative target only relied on color or RGBD images.

\subsection{Deep Q-learning \label{section4_1}}
DQN \cite{mnihHumanlevelControlDeep2015} is one of the most famous value-based reinforcement learning algorithms, which leverages powerful representational ability of deep neural networks to approximate the optimal action value function $Q^{\ast}(s, a)$.

\begin{lemma} \label{lemma1}
	The action value function $Q_{i+1}(s, a)$ can approximate $Q^{\ast}(s, a)$ by iterative update with Bellman equation, when $i \rightarrow \infty$:
	\begin{equation}
		Q_{i+1}(s, a) = \mathbb{E} \left[r_t + \gamma\max_{a'} Q_i(s', a') | s, a\right]
	\end{equation}
\end{lemma}

According to Lemma \ref{lemma1}, DQN can be trained with loss function $\mathcal{L}(\theta_i)$:
\begin{align}
	\mathcal{L}(\theta_i) = \mathbb{E}_{(s, a, r, s') \sim U(D)}\left[(y_i - Q(s, a; \theta_i))^2\right] \label{eq_12}
\end{align}
in which, $y_i = r_t + \gamma\max_{a'} Q(s', a'; \theta^{-}_i)$ is Temperal-Difference (TD) target, $\theta_i$ represents deep Q-network at i-th iteration, and $\theta^{-}_i = \theta_{i//N}$ denotes target network that are updated periodically. The samples $(s, a, r, s')$ utilized to train Q-network are drawn uniformly at random from memory replay pool $D$, also named as experience replay mechanism, which decreases the correlations in observation sequence.

\subsection{Action Space and Reward Function \label{section4_3}}
\begin{figure}
	\centering
	\includegraphics[width=0.3\textwidth]{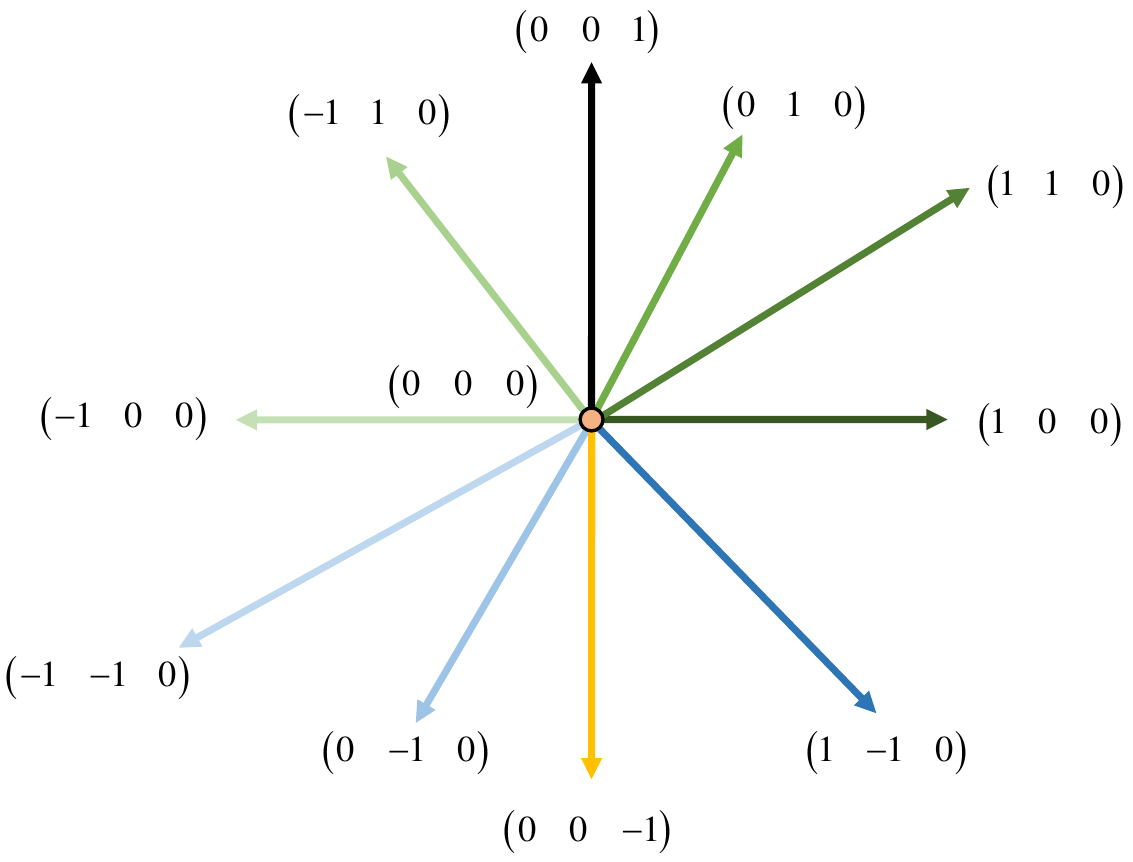}
	\caption{The definition of action space of chaser}
	\label{fig10}
\end{figure}

\newcommand{\tabincell}[2]{\begin{tabular}{@{}#1@{}}#2\end{tabular}}
\begin{table*}[t]
	\centering
	\caption{The evaluation results of PBVS baseline and DRLAVT algorithm. An agent with random policy is also adopted as comparison. The top two places under each metrics are separately highlighted by \textcolor{green}{green} and \textcolor{blue}{blue}.}
	\label{table2}
	\begin{tabular}{ccccccccc}
		\toprule
		\multirow{2}{*}{Name} & \multicolumn{2}{c}{Input} & \multicolumn{3}{c}{Output} & \multicolumn{3}{c}{Metric} \\
		\cline{2-9}
		  & Color & RGBD & Force & Velocity & Position & AEL & AER & Avg Speeds \\
		\midrule

		\multirow{3}{*}{Random} & - & $\surd$ & $\surd$ & - & - & 60.72 & -863.89 & 7194.12 \\
		& - & $\surd$ & - & $\surd$ & - & 28.94 & -969.64 & 6905.11 \\
		& - & $\surd$ & - & - & $\surd$ & 128.40 & -1083.20 & 7230.89 \\
		\hline

		\multirow{2}{*}{\tabincell{c}{PBVS\\(SiamRPN)}} & - & $\surd$ & $\surd$ & - & - & 409.10 & \textcolor{blue}{1182.91} & 76.99 \\
		& - & $\surd$ & - & $\surd$ & - & 679.56 & \textcolor{green}{2790.78} & 74.31 \\
		\hline

		\multirow{2}{*}{\tabincell{c}{PBVS\\(KCF)}} & - & $\surd$ & $\surd$ & - & - & 366.58 & 904.32 & 91.11 \\
		& - & $\surd$ & - & $\surd$ & - & 591.06 & 245.40 & 87.02 \\
		\hline

		\multirow{2}{*}{\tabincell{c}{DRLAVT\\(ConvNet)}} & $\surd$ & - & - & - & $\surd$ & 963.52 & -569.24 & \textcolor{green}{743.68} \\
		& - & $\surd$ & - & - & $\surd$ & \textcolor{green}{1001} & 381.74 & \textcolor{blue}{683.42} \\
		\hline

		\multirow{2}{*}{\tabincell{c}{DRLAVT\\(ResNet-18)}} & $\surd$ & - & - & - & $\surd$ & \textcolor{blue}{987.3} & -410.29 & 293.36 \\
		& - & $\surd$ & - & - & $\surd$ & \textcolor{green}{1001} & 476.07 & 282.68 \\
		\hline

		\multirow{2}{*}{\tabincell{c}{DRLAVT\\(ResNet-34)}} & $\surd$ & - & - & - & $\surd$ & 992.78 & -180.96 & 185.51 \\
		& - & $\surd$ & - & - & $\surd$ & \textcolor{green}{1001} & 530.70 & 179.27 \\
		\hline

		\multirow{2}{*}{\tabincell{c}{DRLAVT\\(ResNet-50)}} & $\surd$ & - & - & - & $\surd$ & \textcolor{green}{1001} & -755.14 & 130.57\\
		& - & $\surd$ & - & - & $\surd$ & \textcolor{green}{1001} & 382.91 & 129.37 \\
		\bottomrule
	\end{tabular}
\end{table*}

At first, reasonable predefined action space in low dimension is necessary for DRLAVT algorithm. We therefore consider a simple action set included 11 types of translational actions in 3D space without rotational operation, which makes DQN convergent quickly during training stage. The details of action space defined in chaser body-fixed coordinate system $\mathcal{F}_B$ can be seen in Fig \ref{fig10}. For example, if take an action like $(1, -1, 0)$, chaser spacecraft should move forward 1 step in x axis and backforward 1 step in y axis. In simulation environment, we assume one step equal to 0.5m which is realised by low-level servoing controller.  

Reward is the source of intelligence of agent in reinforcement learning, like labelled data in supervised learning. What the agent learned from extensive trials-and-errors is heavily depended on the definition of reward function. Therefore, we formulate a reward function for space non-cooperative object active visual tracking, which includes visible reward term and distance penalty term:
\begin{align}
r_t &= r_{vis} - r_{dist} \label{eq_13}
\end{align} 

In our opinion, the first thing that agent must learn is to keep the space non-cooperative target always in the field of view. We therefore add the visible reward term into reward function. Considering the unbalance of visibility of samples, a higher penalty is given for the case that target is out of view. So the final visible reward is formulated as:
\begin{align}
{r_{vis}} &= \left\{ {\begin{array}{*{20}{l}}
	+ 1, \quad \text{in camera view}\\
	- 5, \quad \text{out of camera view}
	\end{array}} \right.
\end{align}

For active visual tracking, the agent should reduce 3D tracking errors defined in Eq. \ref{eq_3}. Therefore, distance penalty term $r_{dist}$ is also added into reward function:
\begin{align}
	r_{dist} =  e =\left\lVert r^{B}_T - r^{\ast}\right\rVert_2
\end{align}

\begin{figure}
	\centering
	\subfloat[Residual Unit 1]{\includegraphics[width=0.24\textwidth]{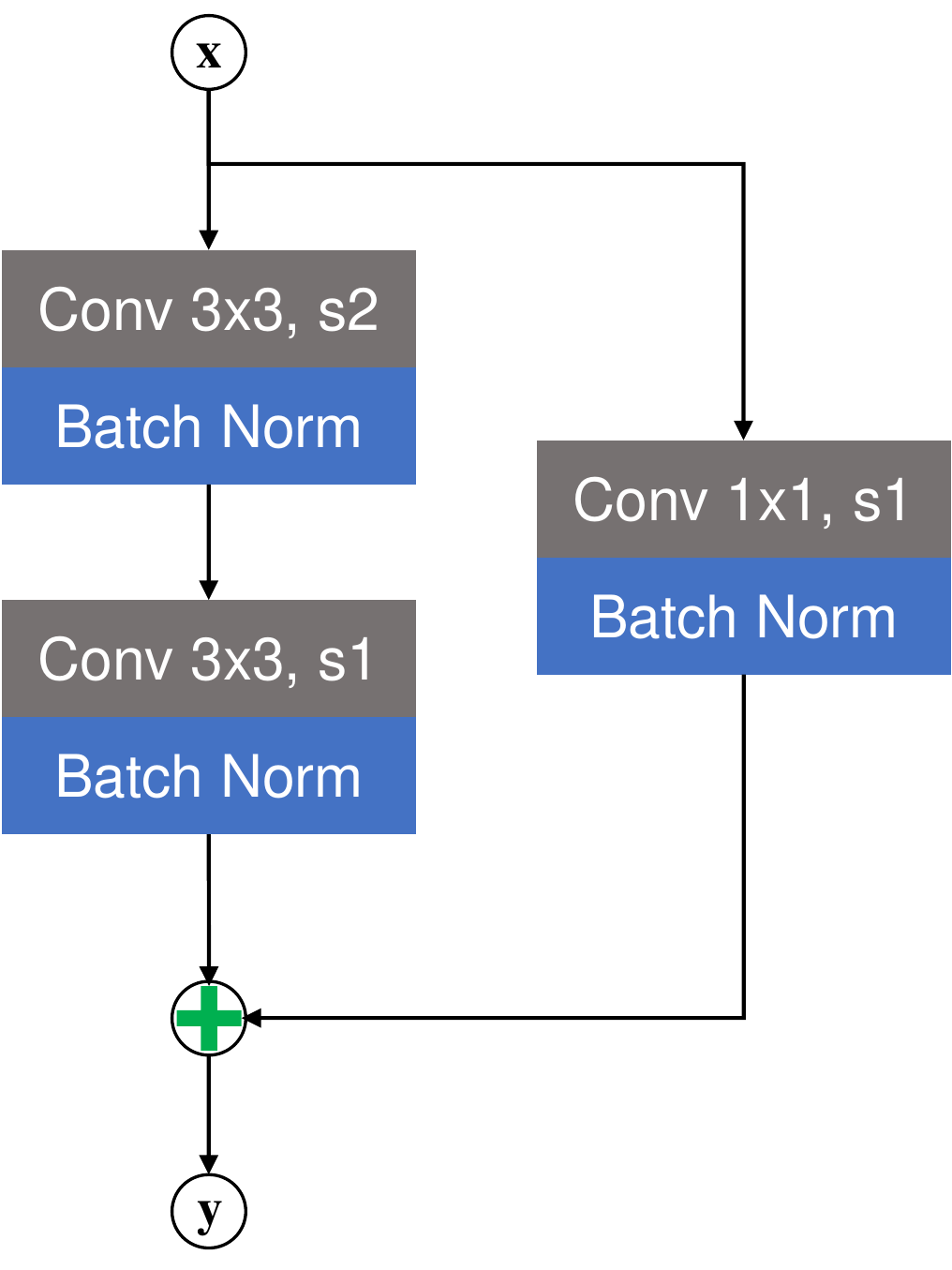}} \hfil
	\subfloat[Residual Unit 2]{\includegraphics[width=0.24\textwidth]{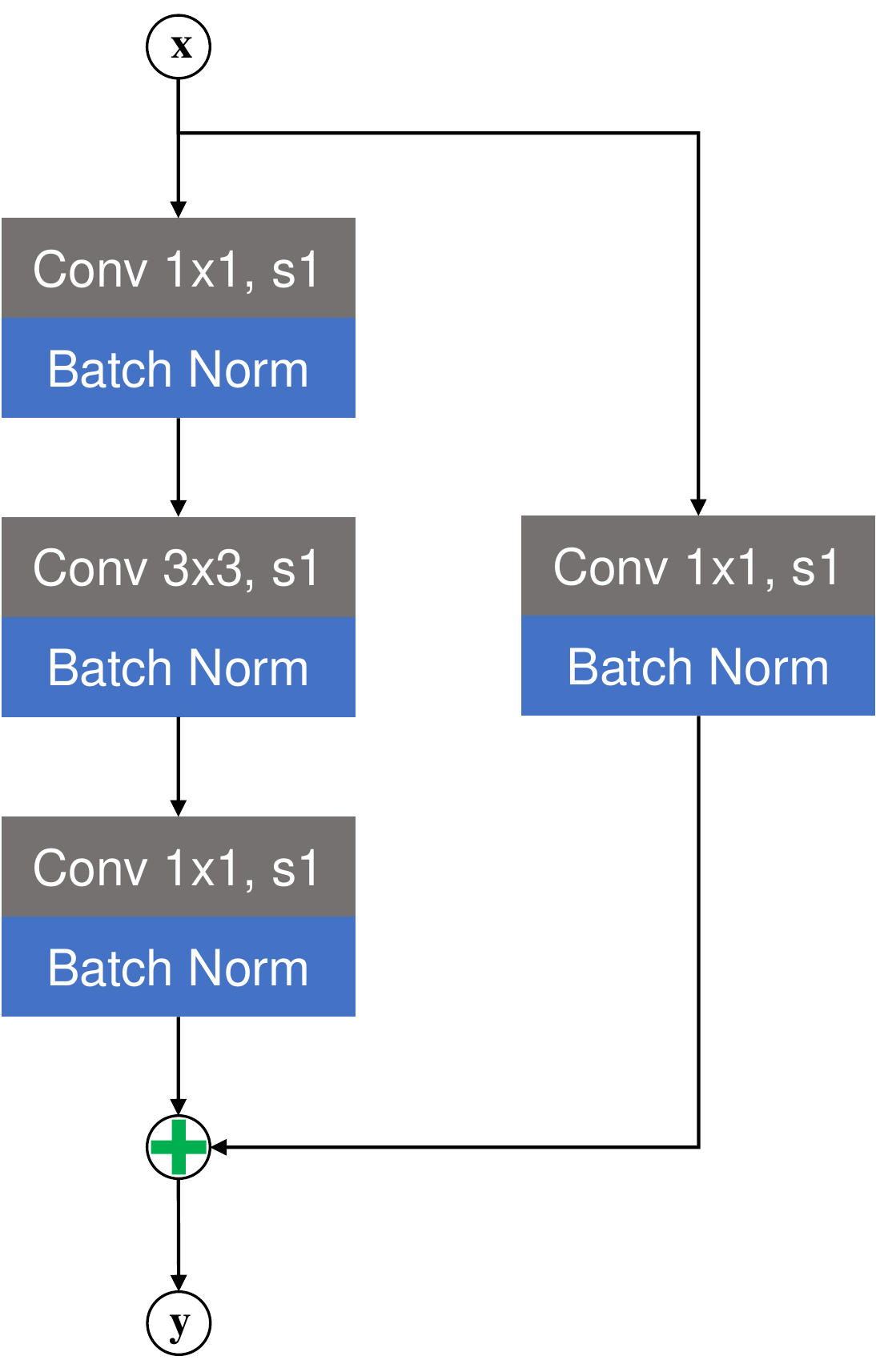}}
	\caption{Two types of residual units. The previous one is used in shallow ResNet, like ResNet-18 and ResNet-34, and the latter one is to consturct deeper ResNet.}
	\label{fig11}
\end{figure}

\begin{figure*}[ht]
	\centering
	\subfloat[active tracking trajectory]{\includegraphics[width=2 in, height= 2 in]{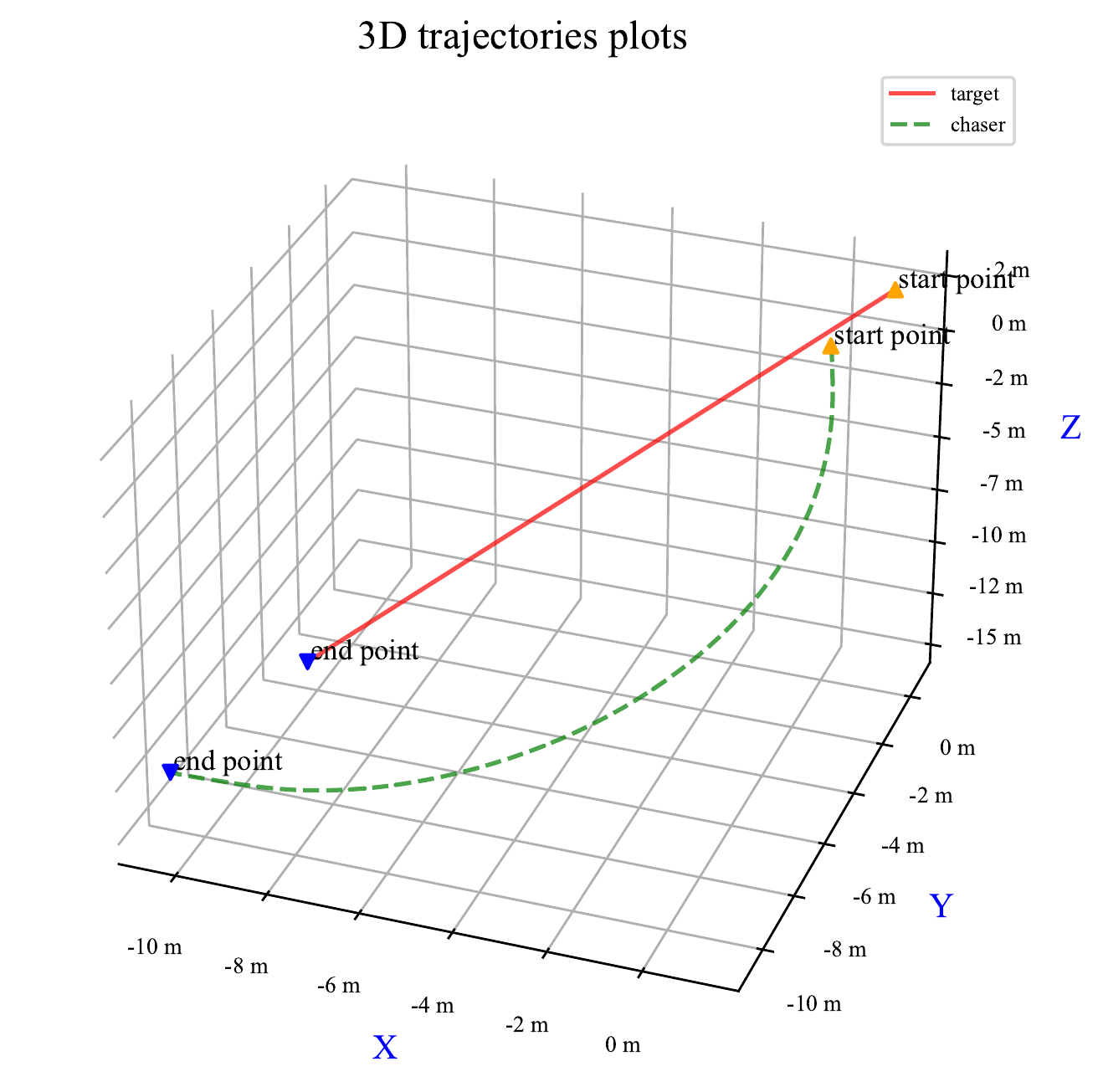}} \hfil
	\subfloat[active tracking errors]{\includegraphics[width= 2.5 in, height= 2 in]{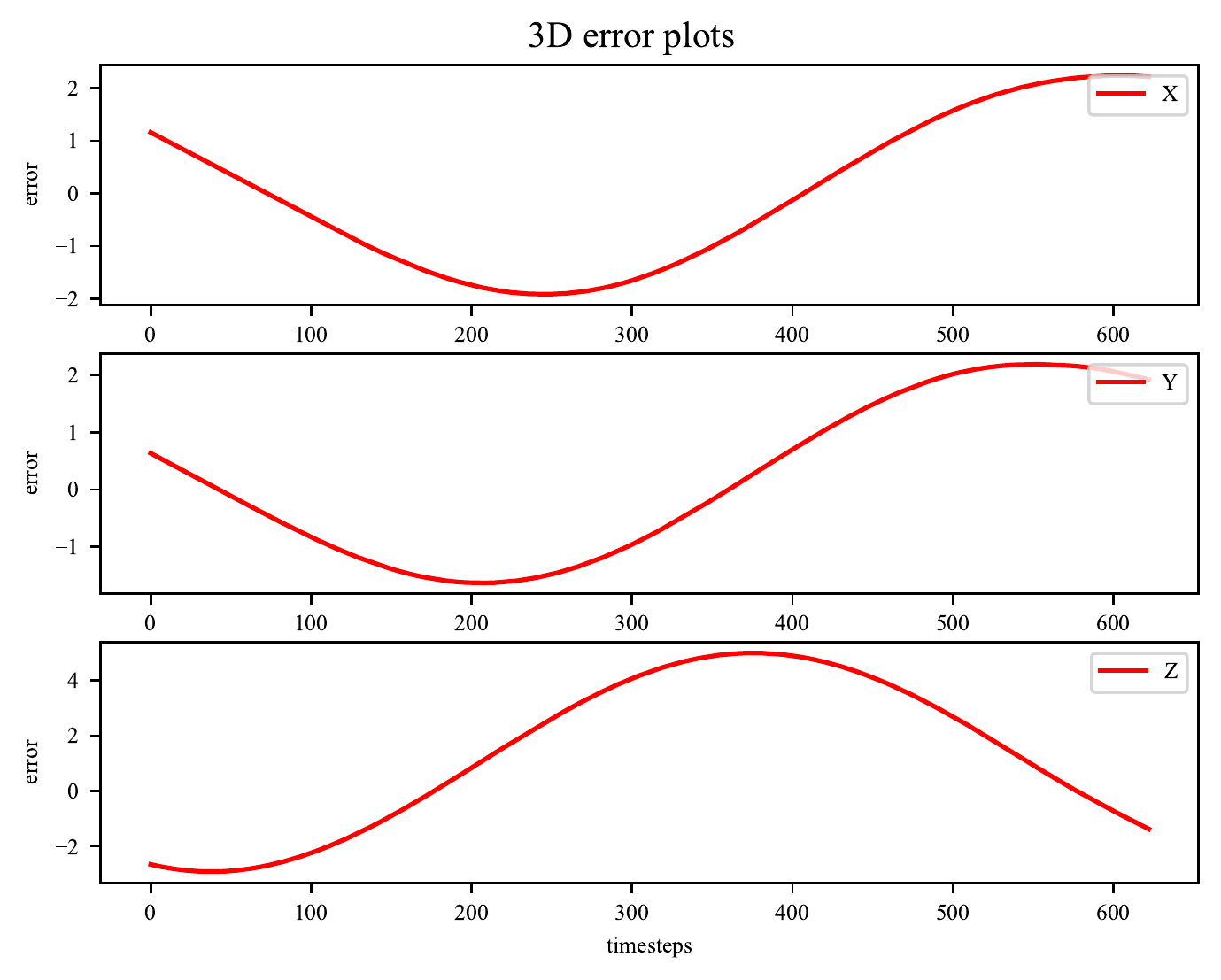}} \vfil
	\subfloat[2D tracking results]{\includegraphics[width=0.9\textwidth]{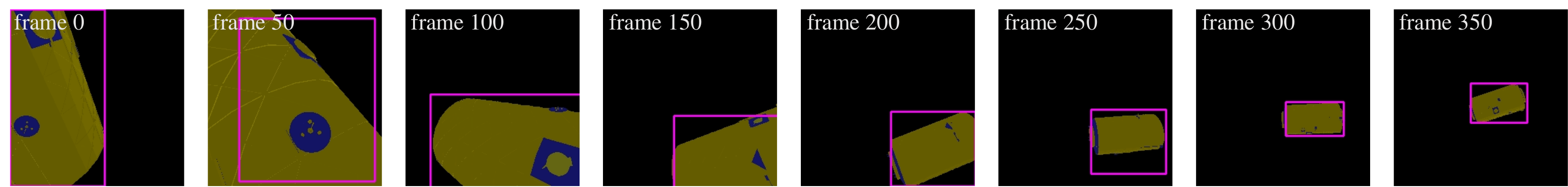}}
	\caption{The results of PBVS based on SiamRPN.}
	\label{fig12}
\end{figure*}

\begin{figure*}[ht]
	\centering
	\subfloat[active tracking trajectory]{\includegraphics[width = 2 in, height = 2 in]{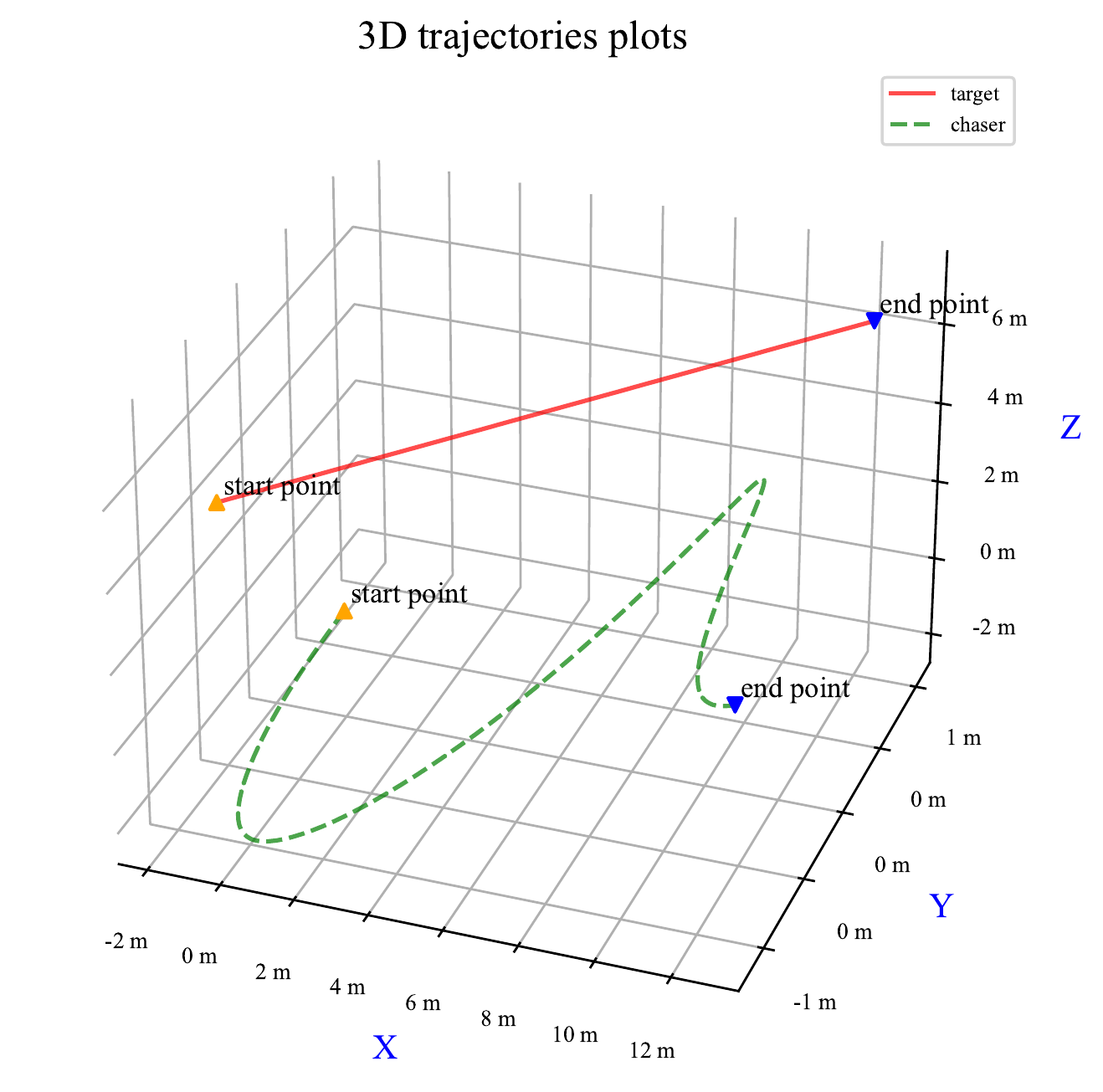}} \hfil
	\subfloat[active tracking errors]{\includegraphics[width= 2.5 in, height= 2 in]{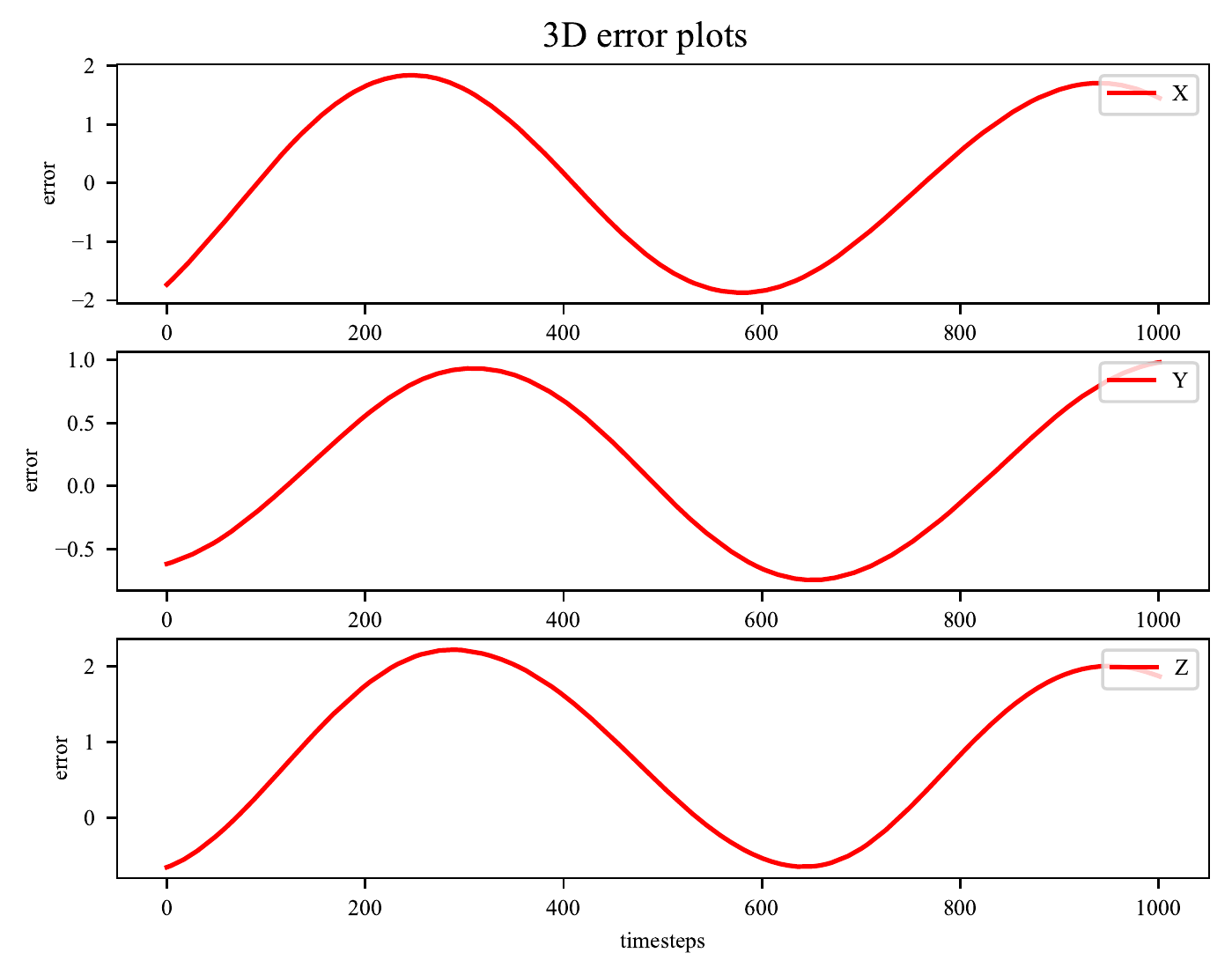}} \vfil
	\subfloat[2D tracking results]{\includegraphics[width=0.9 \textwidth]{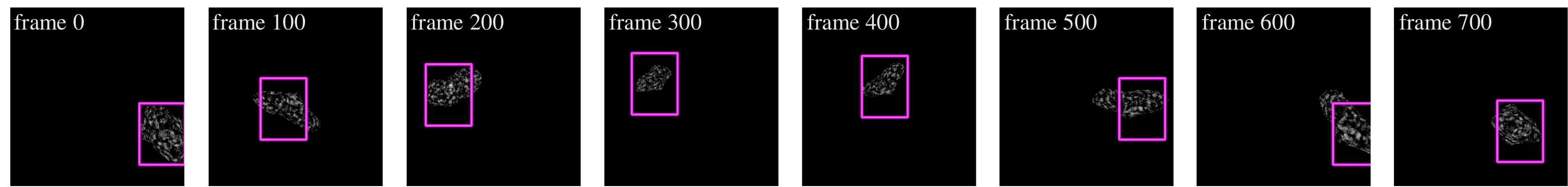}}
	\caption{The results of PBVS based KCF.}
\end{figure*}

\subsection{Q-network Architecture\label{section4_2}}
The core of Deep Q-learning algorithm is to predict action values with observed image by Q-network. In original paper, Mnih \cite{mnihHumanlevelControlDeep2015} utilized a shallow neural network which only consists of three convolutional layers and two full-connected layers. Considering the powerful representational ability of deeper convolutional networks \cite{krizhevsky2017imagenet, he2016deep}, we adopt two types of neural network architectures in this work, which are clearly shown in Fig \ref{fig9}. 

The first architecture proposed by us is ConvNet derived from original DQN. The main improvements include (1) additional convolutional layer with kernel size 1 and stride 1, which aims to merge the features in depth axis of tensor, especially for RGBD images. (2) more max-pooling layer following each convolutional layers, which significantly reduces the dimension of tensor and learns high-level patterns from raw input image. (3) additional full-connected layer and Dropout regularization method. Although colorful tricks are used in ConvNet, it remains tiny and elegant. 

To study the impact of convolutional layers depth on active visual tracking task, we further utilize ResNet \cite{he2016deep} as the second architecture of Q-network here, which can even involve hundreds of convolutional layers without performance degradation due to the property of residual units. Therefore, we develop three ResNets (i.e. ResNet-18, ResNet-34, and ResNet-50) with different configurations as DQN.

\section{Experiments \label{section5}}
In this section, we first evaluate PBVS baseline and DRLAVT algorithms on simulated environment with 20 repetitions, which demonstrates the effective and advancement of our method. The influences of disturbations and reward functions on DRLAVT performance are futher studied. In addition, we also validate DRLAVT with multiple motion patterns to explore what it learnt from hundreds of trial-and-errors. It is worthwhile noting that all the experiments are implemented on HPC server with Intel Xeon@E502650v4@2.2GHz CPU and Nvidia Tesla P100 GPU. 

\begin{figure}
	\centering
	\subfloat[episode length]{\includegraphics[width=0.3\textwidth]{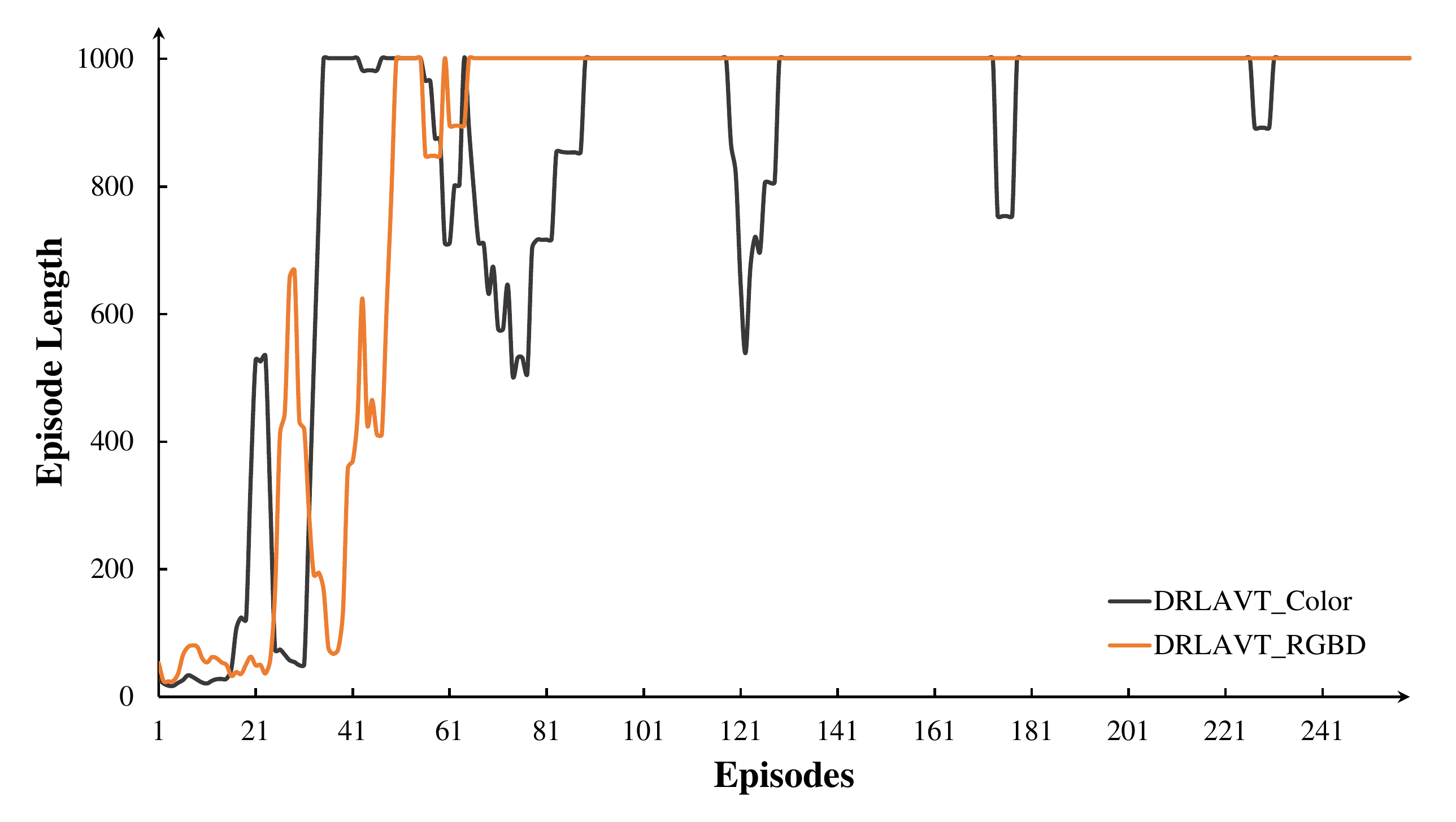}} \vfil
	\subfloat[action values]{\includegraphics[width=0.3\textwidth]{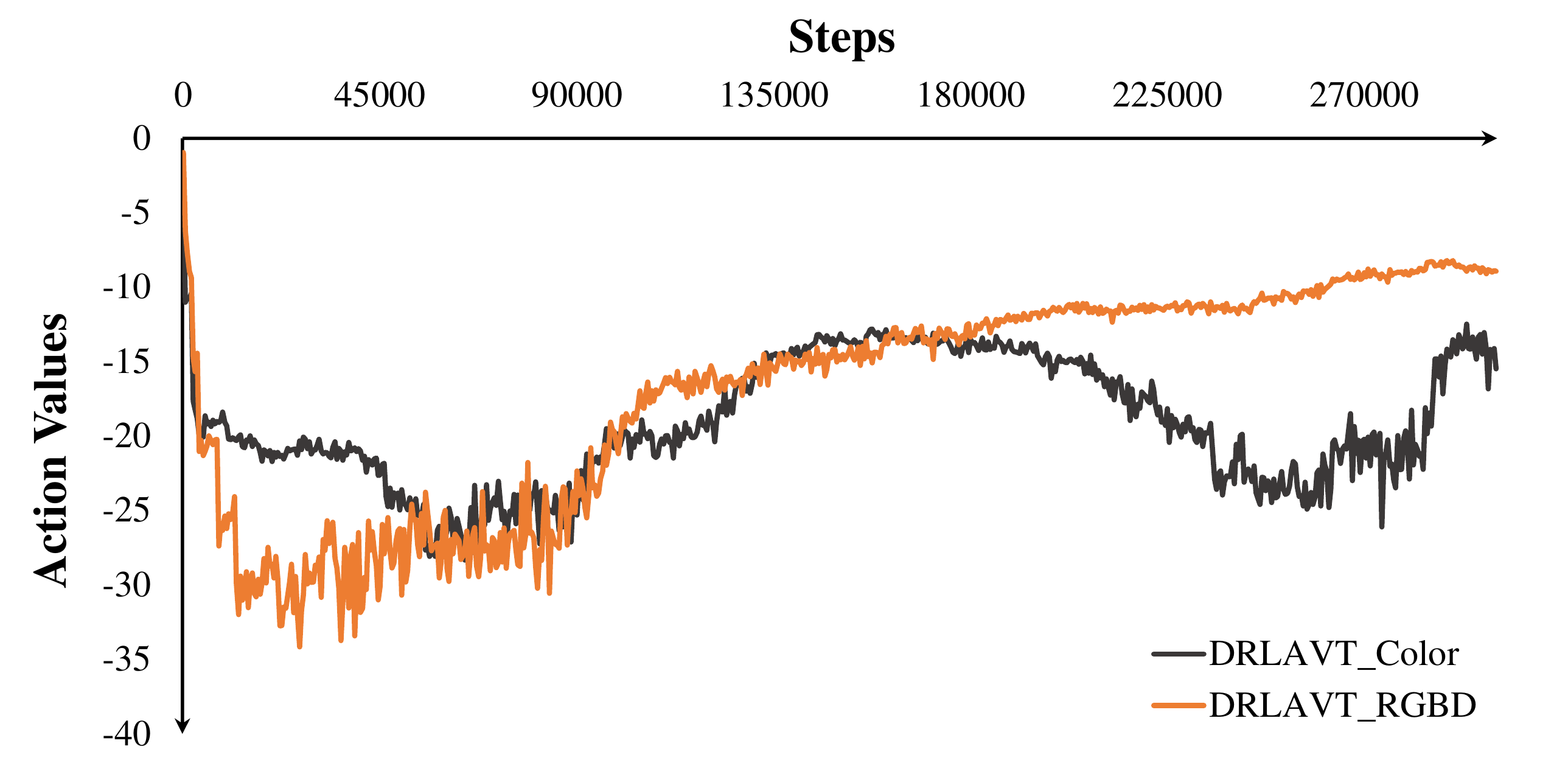}}
	\caption{Training curves of DRLAVT based on ConvNet.}
	\label{fig14}
\end{figure}

\begin{figure*}[ht]
	\centering
	\subfloat[active tracking trajectory]{\includegraphics[width= 2 in, height= 2 in]{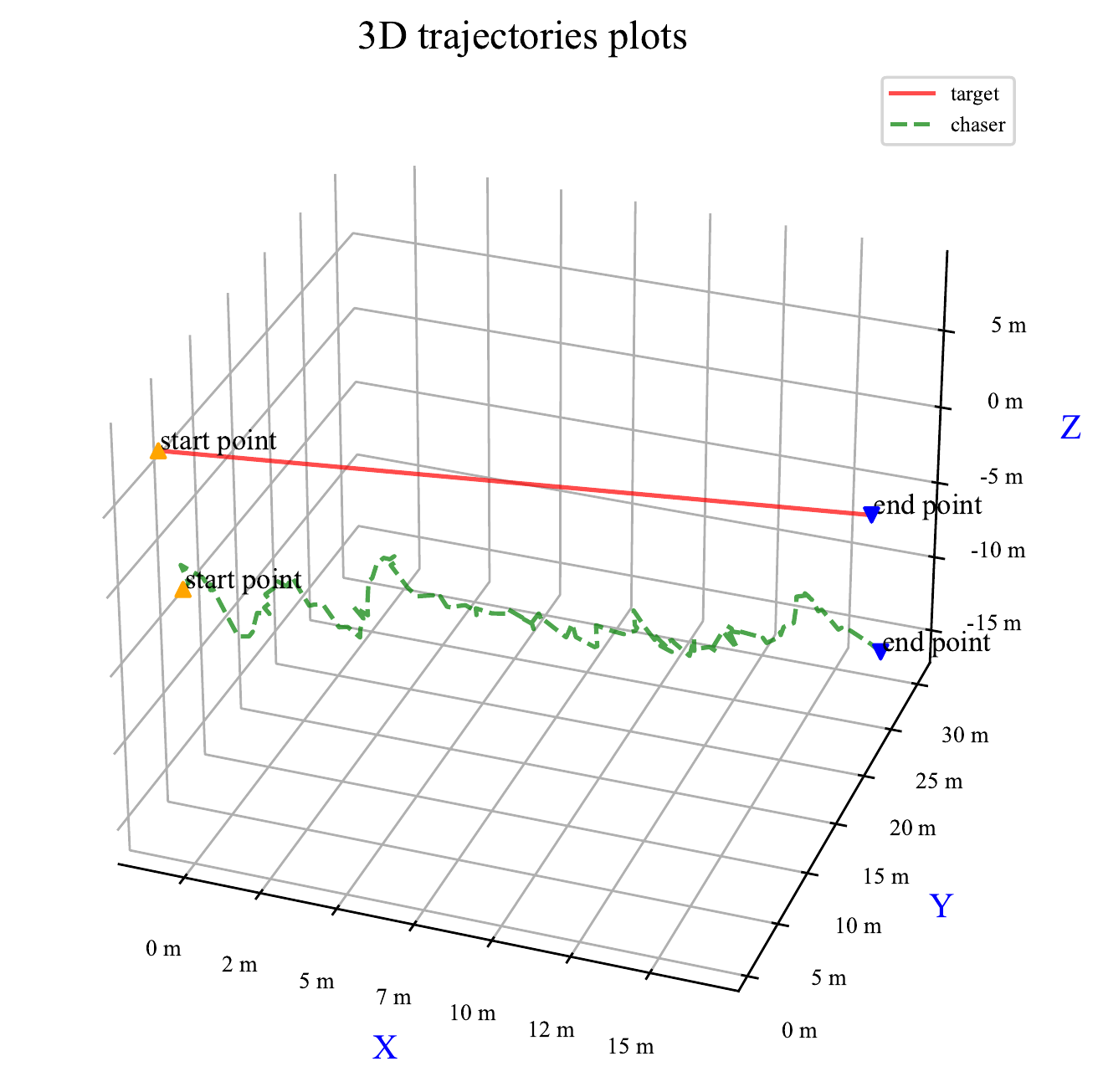}} \hfil
	\subfloat[active tracking errors]{\includegraphics[width= 2.5 in, height= 2 in]{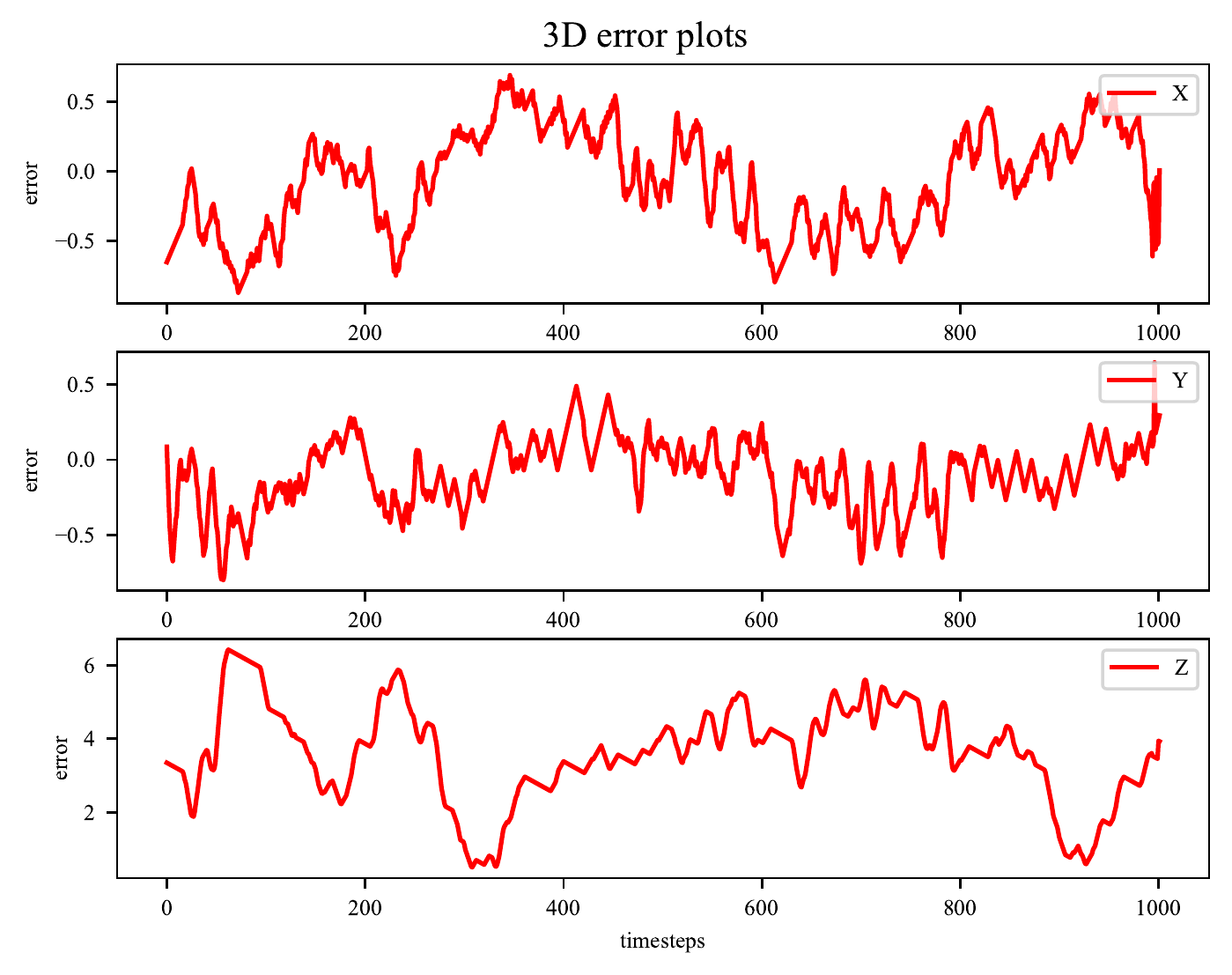}} \vfil
	\subfloat[observed images]{\includegraphics[width=0.9 \textwidth]{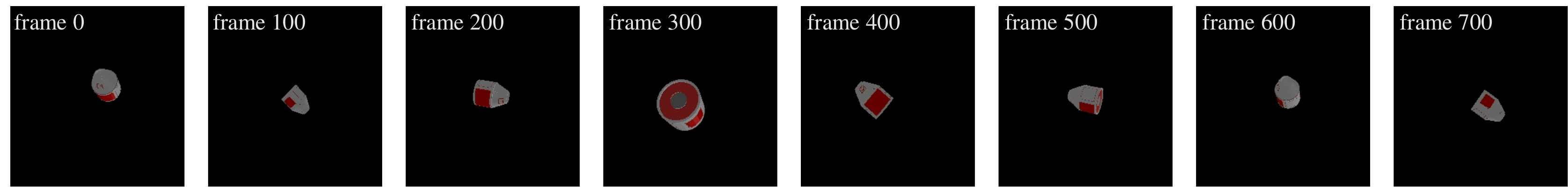}}
	\caption{The results of DRLAVT based on ConvNet with Color image.}
\end{figure*}

\begin{figure*}[ht]
	\centering
	\subfloat[active tracking trajectory]{\includegraphics[width = 2 in, height= 2 in]{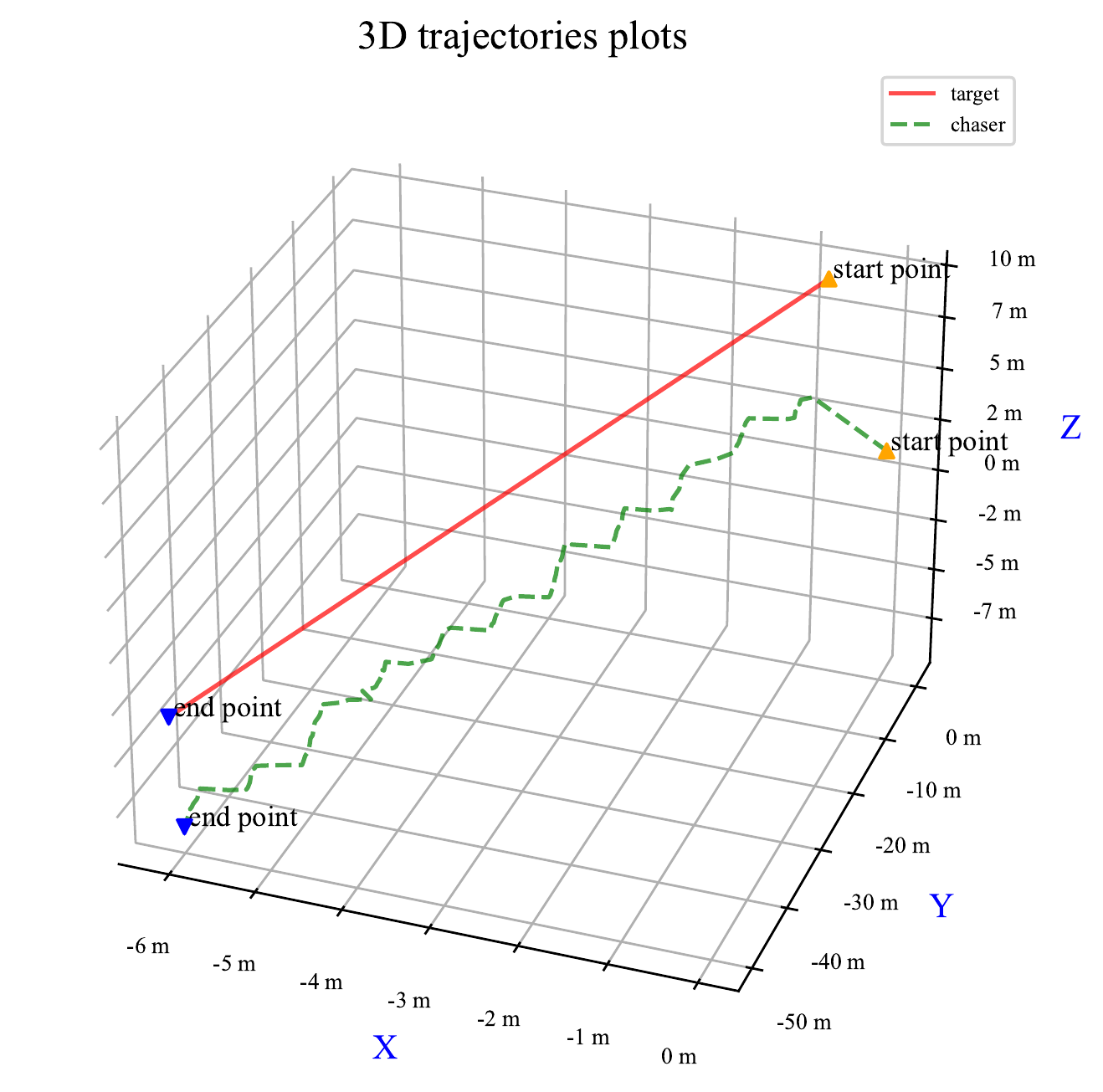} \label{fig13_a}} \hfil
	\subfloat[active tracking errors]{\includegraphics[width= 2.5 in, height= 2 in]{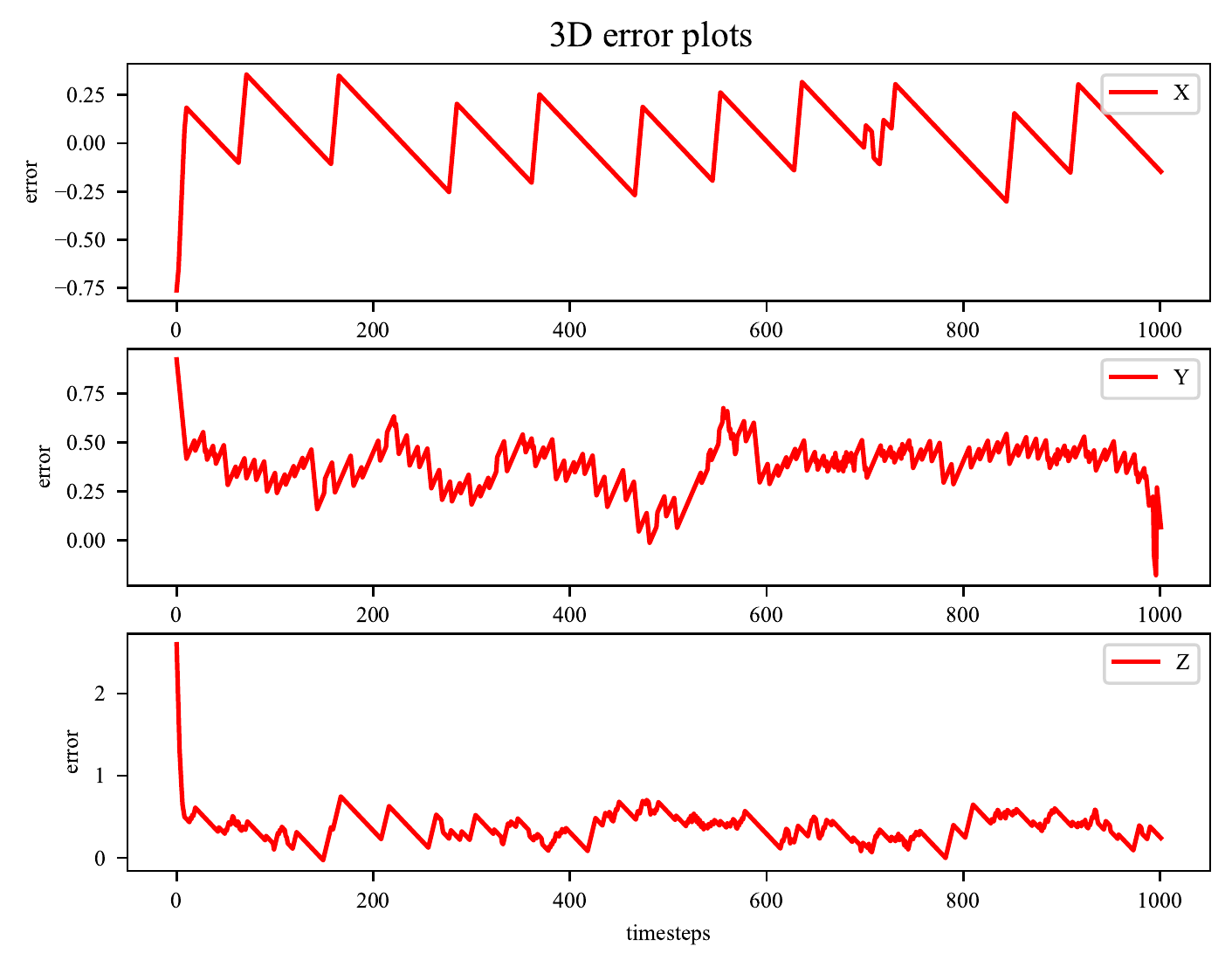} \label{fig13_b}} \vfil
	\subfloat[observed images]{\includegraphics[width=0.9\textwidth]{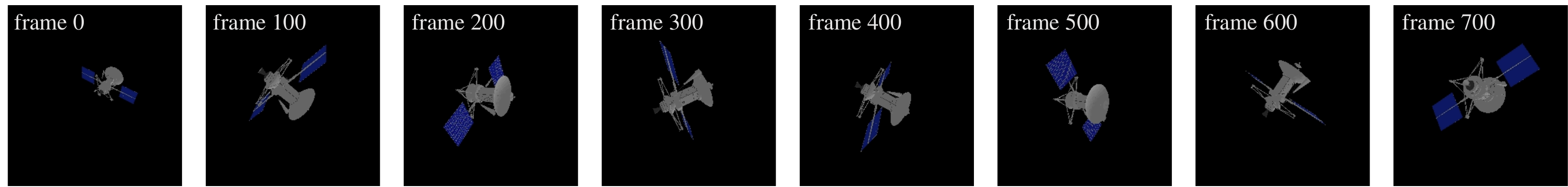} \label{fig13_c}}
	\caption{The results of DRLAVT based on ConvNet with RGBD image.}
	\label{fig13}
\end{figure*}

\subsection{Evaluation Results \label{section5_1}}
\subsubsection{PBVS Baseline}
At first, We evaluate PBVS baseline algorithm in simulated environment, where the initial state of 2D monocular tracker is automatically generated by evaluation toolkit and re-initialization is prohibited if tracker fails. The active tracking results of PBVS baseline can be clearly seen in Fig \ref{fig12}. We find out that the PBVS algorithm based on SiamRPN merely achieves commonplace performance (about 409.10 AEL and 1182.92 AER with 76.99 Hz) and the tracking errors are always in fluctuation, which are caused by the saturation of PID controller, failures of 2D monocular tracker, and inaccurate estimation of 3D position when relative motion occurs between chaser and non-cooperative target. 

We also replace the state-of-the-art SiamRPN tracker in PBVS baseline with classical KCF \cite{henriquesHighSpeedTrackingKernelized2015} tracker that can simply run without GPU. The evaluation results of PBVS baseline based on KCF is listed in Table \ref{table2}. It clearly shows that active tracking performance significantly degrades 23.6\% under AER metric after the replacement of 2D monocular tracker, although there is a slight improvement on running speed with KCF.

Furthermore, we control the velocity of chaser spacecraft in PBVS baseline framework, rather than directly force control. The kinematic model of chaser spacecraft is formulated as:
\begin{align}
	\dot{X}^{I}_{s} = u' + \Omega'
\end{align}
in which, $X^{I}_{s}$ is the position of chaser, $u'$ is the control command of chaser velocity, and $\Omega'$ is noise term. We also assume that $\left\lVert u'\right\rVert _{\infty} \leq 5$. The evaluation results in 5th and 7th rows of Table \ref{table2} demonstrates PBVS baseline with velocity control reach higher performance comparing to orginal algorithm.

\begin{table}
	\centering
	\caption{The training configurations of DRLAVT.}
	\label{table4}
	\begin{tabular}{lcl}
		\toprule
		Params & Value & Note \\
		\midrule
		replay buffer & 50000 & \tabincell{l}{The size of replay pool }\\
		initial buffer & 10000 & \tabincell{l}{The number of initial \\ experiences} \\
		episode num & 300 & \tabincell{l}{The number of episodes \\ used to train Q-network}\\
		max episode len & 1000 & \tabincell{l}{The max length of one \\ episode, but if target is \\ lost, episode will be over}\\
		update interval & 10 & \tabincell{l}{The update interval of \\ target network }\\
		gamma & 0.99 & \tabincell{l}{Rewards discount factor}\\
		\bottomrule
	\end{tabular}
\end{table}

\begin{table*}[t]
	\centering
	\caption{The impact of different perturbations on active visual tracking algorithms.}
	\label{table3}
	\begin{tabular}{ccccccccc}
		\toprule
		\multirow{3}{*}{Name} & \multicolumn{6}{c}{Perturbations} & \multicolumn{2}{c}{Metric} \\
		\cline{2-9}
		& \multirow{2}{*}{\tabincell{c}{Actuator \\ noise}} & \multirow{2}{*}{\tabincell{c}{Time \\ delay}} & \multicolumn{4}{c}{Image blur} & \multirow{2}{*}{AEL} & \multirow{2}{*}{AER} \\ 
		\cline{4-7}
		& & & level 1 & level 2 & level 3 & level 4 & & \\
		\midrule
		\multirow{7}{*}{\tabincell{c}{PBVS\\(SiamRPN)}} 
		 & $\surd$ &  & - & - & - & - & 220.18 & 39.73 \\
		 & - & $\surd$ & - & - & - & - & 423.67 & 1186.23 \\
		 & - & - & $\surd$ & - & - & - & 409.85 & 1182.91 \\
		 & - & - & - & $\surd$ & - & - & 390.48 & 1050.84 \\
		 & - & - & - & - & $\surd$ & - & 402.52 & 1108.77 \\
		 & - & - & - & - & - & $\surd$ & 393.74 & 1072.30 \\
		 & $\surd$ & $\surd$ & - & - & $\surd$ & - & 189.41 & -83.41 \\
		 \hline
		 \multirow{7}{*}{\tabincell{c}{DRLAVT\\(ConvNet)}}
		 & $\surd$ &  & - & - & - & - & 985.54 & -89.90 \\
		 & - & $\surd$ & - & - & - & - & 997.04 & -36.80 \\
		 & - & - & $\surd$ & - & - & - & 1001 & 211.26 \\
		 & - & - & - & $\surd$ & - & - & 1001 & 112.13 \\
		 & - & - & - & - & $\surd$ & - & 1001 & 35.85 \\
		 & - & - & - & - & - & $\surd$ & 1001 & -37.66 \\
		 & $\surd$ & $\surd$ & - & - & $\surd$ & - & 983.50 & -60.78 \\
		\bottomrule
	\end{tabular}
\end{table*}

\subsubsection{DRLAVT}
The DRLAVT based on ConvNet that directly adopts color image to select action is evaluated firstly, of which results are listed at 8th row of Table \ref{table2}. Our DRLAVT greatly outperforms PBVS baseline algorithm under AEL metric and runs up to 743.68 Hz without tracking initialization. Although it achieves worse AER score comparing to baseline algorithm, because of the difficulty for agent with reinforcement learning to learn depth information from color image. To this end, we provide color image along with extra depth map to DRLAVT and re-train it from scratch. The training curves of DRLAVT based on ConvNet are depicted Fig \ref{fig14}. Note that the depth map before fetched into network has been normalized to $\left[0, 1\right]$ by following equation:
\begin{align}
	I_{depth} = \frac{I_{depth}}{z_{\max}^C}
\end{align}
in which, $z_{\max}^C$ is the maximumn distance that simulated vision sensors can perceive. As we expected, DRLAVT with RGBD image reach much higher performance, which can also be seen from the active tracking trajectory in Fig \ref{fig13_a} and tracking errors in Fig \ref{fig13_b}. Chaser can quickly approach to the target at the start of tracking and reduce tracking errors to 0 with small oscillation under the guidance of DRLAVT. 

In addition, DRLAVT that utilizes different ResNets as Q-network are also evaluated, including ResNet-18, ResNet34, and ResNet-50. The results summarized in the final 6 rows of Table \ref{table2} clearly shows ResNet can further improve the performance of DRLAVT, because of its large model capacity and powerful representational ability. Especially, DRLAVT with ResNet-34 can achieve 1001 AEL and 530.70 AER. However, performance degradation inevitably occurs with the depth of ResNet increasing.

\subsubsection{Impact of Noise}
The robustness of active visual trackers is of significance in application. Therefore, we evaluate both PBVS baseline and DRLAVT algorithm under different types of perturbations, such as actuator noise, processing time-delay, and image blur. The experiment results are listed in Table \ref{table3}. We find out that actuator noise and time delay severely decrease the AEL and AER metrics of both active trackers. Meanwhile, it is evident that DRLAVT is more sensitive to image blur than PBVS algorithm which benefits from the complexity architecture of SiamRPN, as blurring level increasing. Although the influence of image blur is relative slight comparing to other perturbations. 

On the whole, however, our DRLAVT algorithm is more robuster to different disturbations, of which AEL metric only degrades 1.7\% even in the presence of three perturbations together.  

\begin{table}
	\centering
	\caption{The evaluation results of DRLAVT that is trained with multiple targets or single target.}
	\label{table5}
	\begin{tabular}{cccc}
		\toprule
		Name & Targets & AEL & AER \\
		\midrule
		\multirow{6}{*}{\tabincell{c}{DRLAVT \\ (Multiple)}} & Asteroid 06 & 1001 & -366.32 \\
		& Capsule 03 & 1001 & 552.26 \\
		& Rocket 03 & 1001 & 579.14 \\
		& Satellite 03 & 1001 & 653.98 \\
		& Station 03 & 1001 & 489.66 \\
		\cline{2-4}
		& \textcolor{blue}{overall} & 1001 & 381.74 \\
		\hline
		\multirow{6}{*}{\tabincell{c}{DRLAVT \\ (Single)}} & Asteroid 06 & 1001 & -218.06 \\ 
		& Capsule 03 & 1001 &  275.92 \\
		& Rocket 03 & 924 &  -2696.51\\
		& Satellite 03 & 903 & -2756.19 \\
		& Station 03 & 891 & -3155.57 \\
		\cline{2-4}
		& \textcolor{blue}{overall} & 944 & -1710.08 \\
		\bottomrule
	\end{tabular}
\end{table}

\subsection{Further Study on DRLAVT \label{section5_2}}
\subsubsection{Multiple Targets}
Paper \cite{loquercioDeepDroneRacing2020} presented that extensive simulated data generated by domain randomization method could make agent more robust to environment variations and reach zero-shot simulation-to-real transfer on the task of drone racing. To this end, our DRLAVT is also trained about 300 episodes with 12 space non-cooperative objects, which aims to make agent to learn various motion patterns of targets and optimal guidance policy, rather than specific geometries or textures. As comparison, an extreme case is introduced that only one target (Asteroid 01) is used to train DRLAVT agent with the same configurations. 

The experiments are summarized in Table \ref{table5}, which include detailed evaluation measurements for 5 type of unseen targets (i.e. Asteroid06, Capsule03, Rocket03, Satellite03, Station03). It clearly shows that DRLAVT trained with single target only achieves comparable results on similar targets, such as Asteroid06 and Capsule03. However, DRLAVT trained by multiple targets can successfully track all the targets with high performance.

\subsubsection{Reward Functions}

\begin{figure}
	\centering
	\subfloat[episode length]{\includegraphics[width=0.3\textwidth]{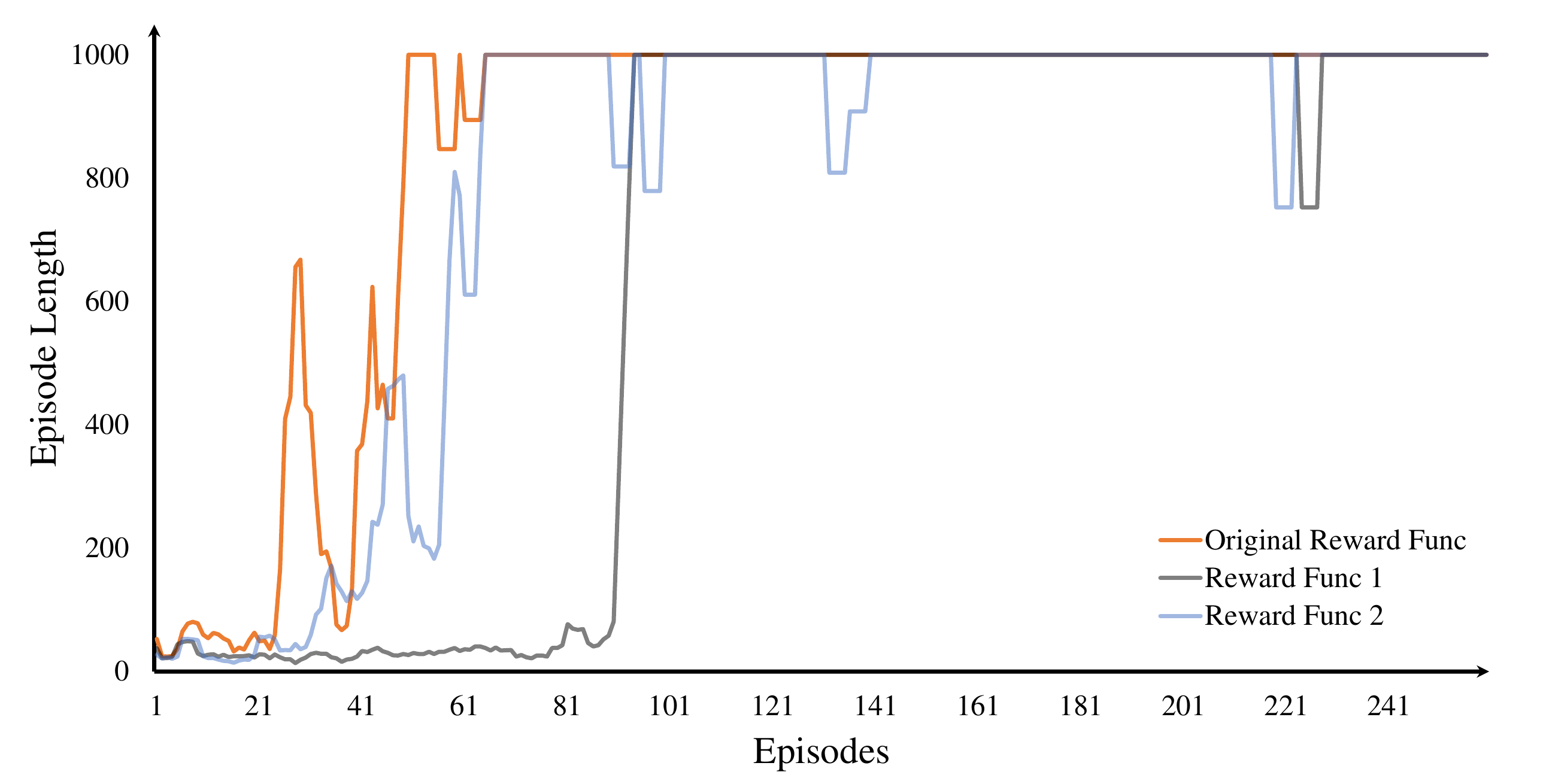}} \vfil
	\subfloat[action values]{\includegraphics[width=0.3\textwidth]{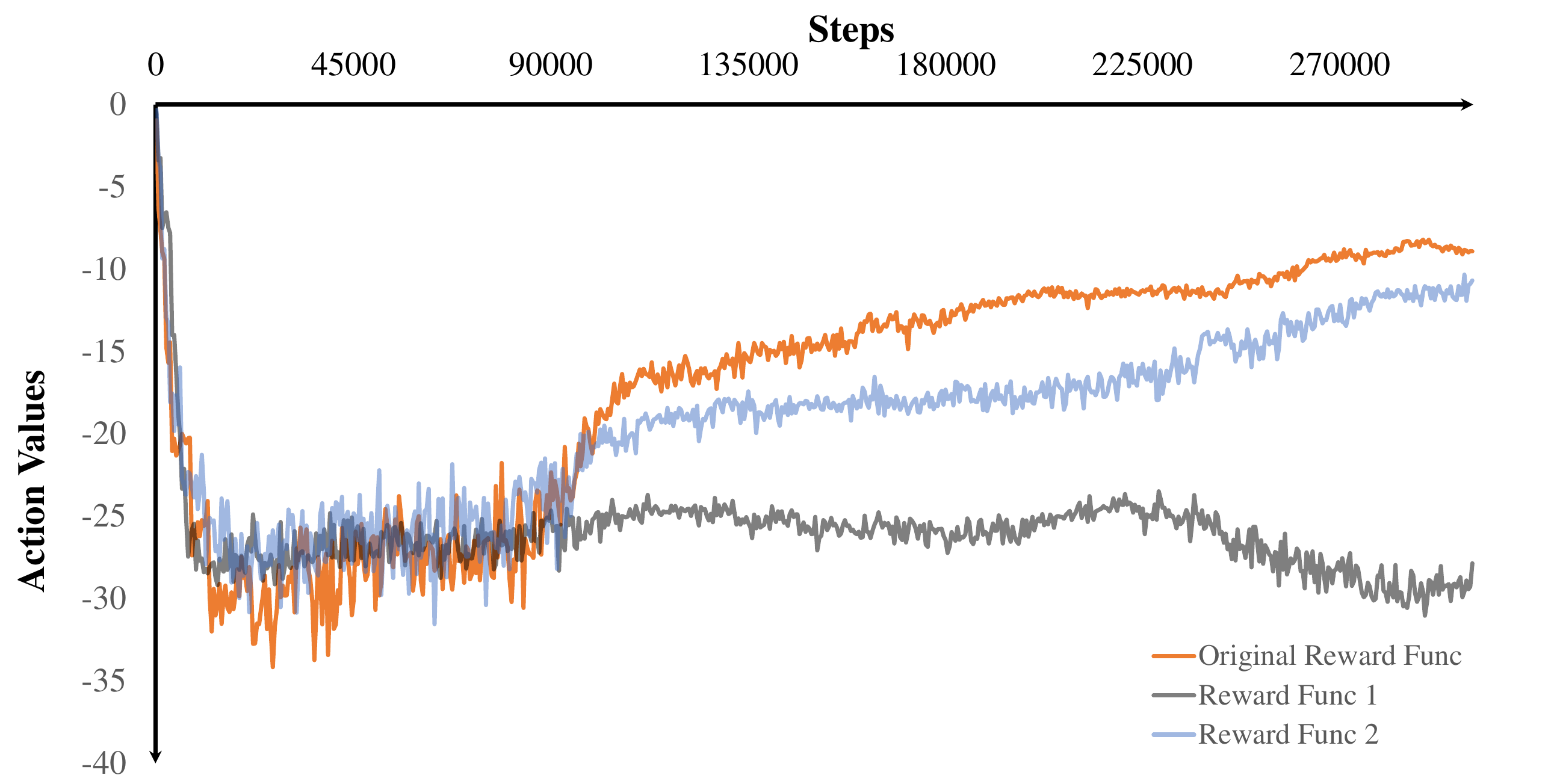}}
	\caption{Training curves of DRLAVT with different reward functions.}
	\label{fig15}
\end{figure}

\begin{table}
	\centering
	\caption{The peformance of DRLAVT with different reward functions.}
	\label{table6}
	\begin{tabular}{ccc}
		\toprule
		Reward Function & AEL & AER \\
		\midrule
		original & 1001 & 391.74 \\
		reward func 1 & 786.08 & -1193.15\\
		reward func 2 & 1001 & -376.93\\
		\bottomrule
	\end{tabular}
\end{table}

We carry out abalation study on reward function defined in Eq. \ref{eq_13} to explore how it impacts on the performance of DRLAVT. At first, we remove the visible term in original reward function:
\begin{align}
	r^1_t &= r_{dist} \label{eq_18}
\end{align}

In addition, we modify the penalty in visible term when the target is out of view, which is formulated as follows:
\begin{align}
	{r'_{vis}} &= \left\{ {\begin{array}{*{20}{l}}
		+ 1, \quad \text{in camera view}\\
		- 1, \quad \text{out of camera view}
		\end{array}} \right.
\end{align}
Therefore, we define the second new reward funciton:
\begin{align}
	r^2_t = r'_{vis} + r_{dist} \label{eq_20}
\end{align}

The evaluation results of DRLAVT retrained from scratch by using Eq. \ref{eq_18} and \ref{eq_20} are listed in Table \ref{table6}. We also depict the episode length and action values curves during training in Fig \ref{fig15} which clearly proves that the visible term is of significance to accelerate the training period of DRLAVT. In addition, the higher penalty in visible terms can also improve the action values that DRLAVT predicts, as we expected. 

\begin{figure}
	\centering
	\begin{minipage}[b]{0.48\textwidth}
		\subfloat[Satellite03 with left translation]{\includegraphics[width=1in, height=0.7in]{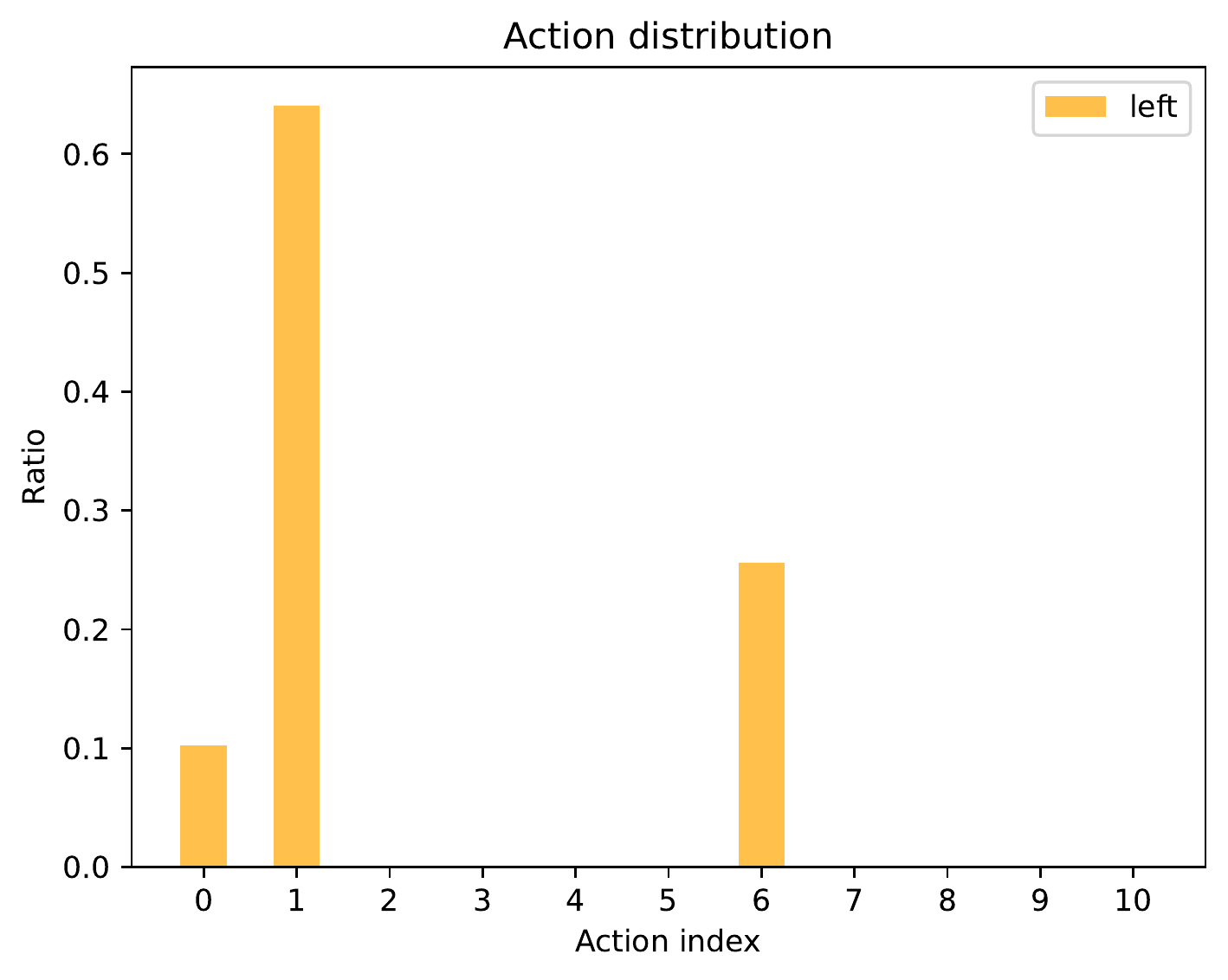} 
		\includegraphics[width=2.1in,height=0.7in]{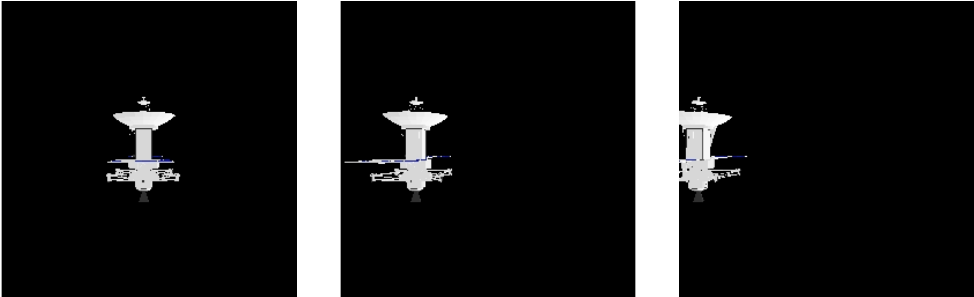} \label{fig16_1}} 

		\subfloat[Satellite03 with right translation]{\includegraphics[width=1in, height=0.7in]{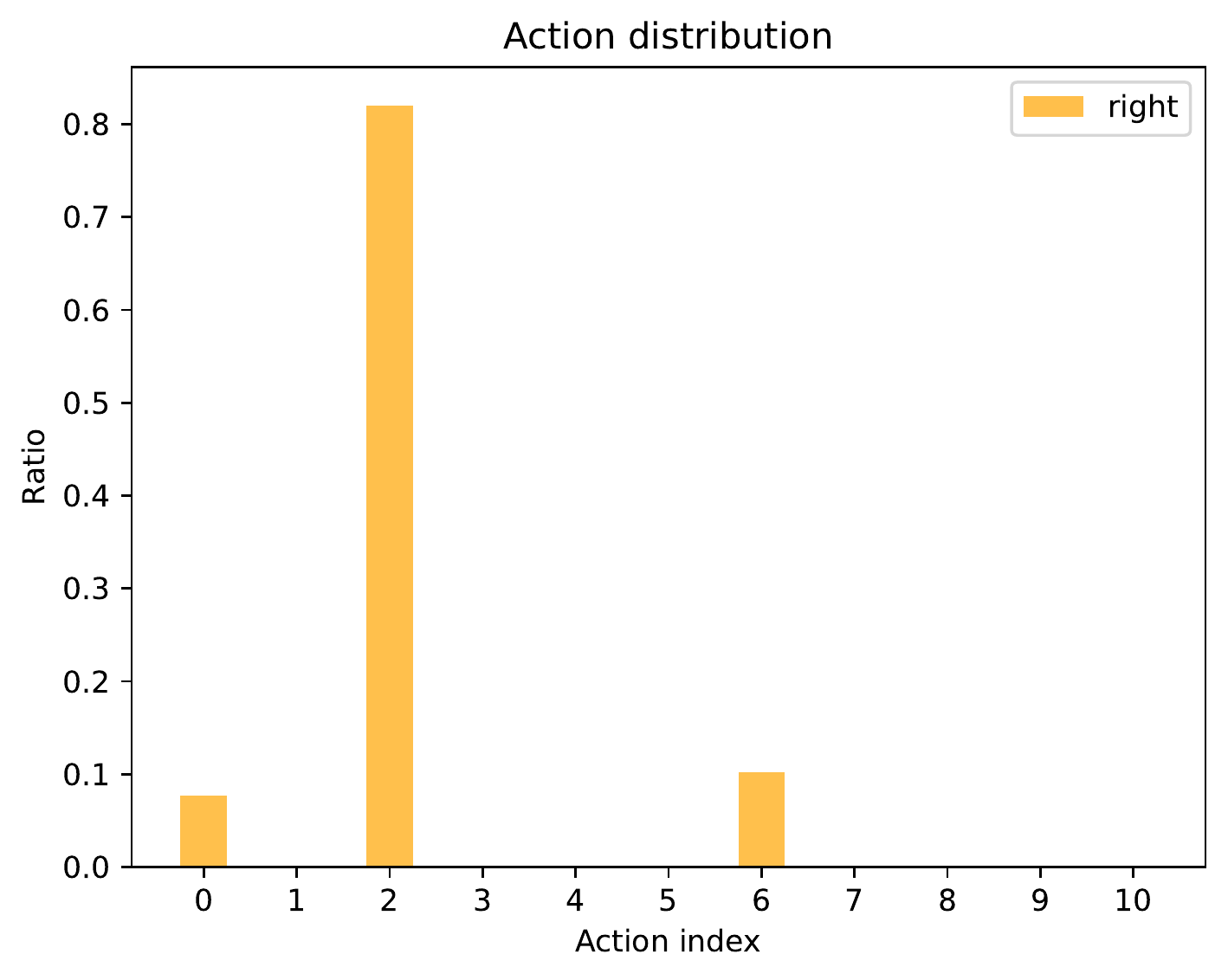} 
		\includegraphics[width=2.1in,height=0.7in]{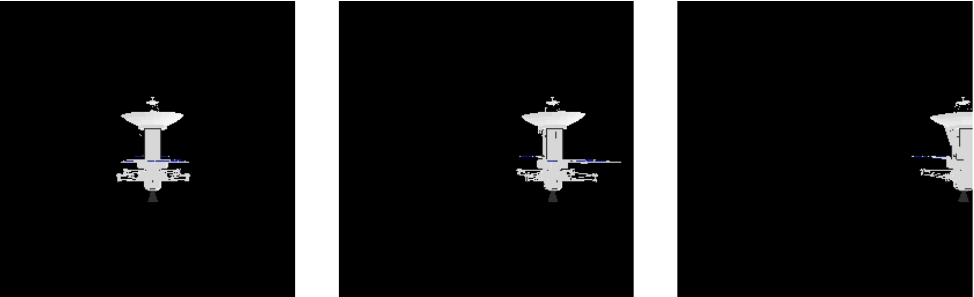} \label{fig16_2}} 

		\subfloat[Asteroid06 with left translation]{\includegraphics[width=1in, height=0.7in]{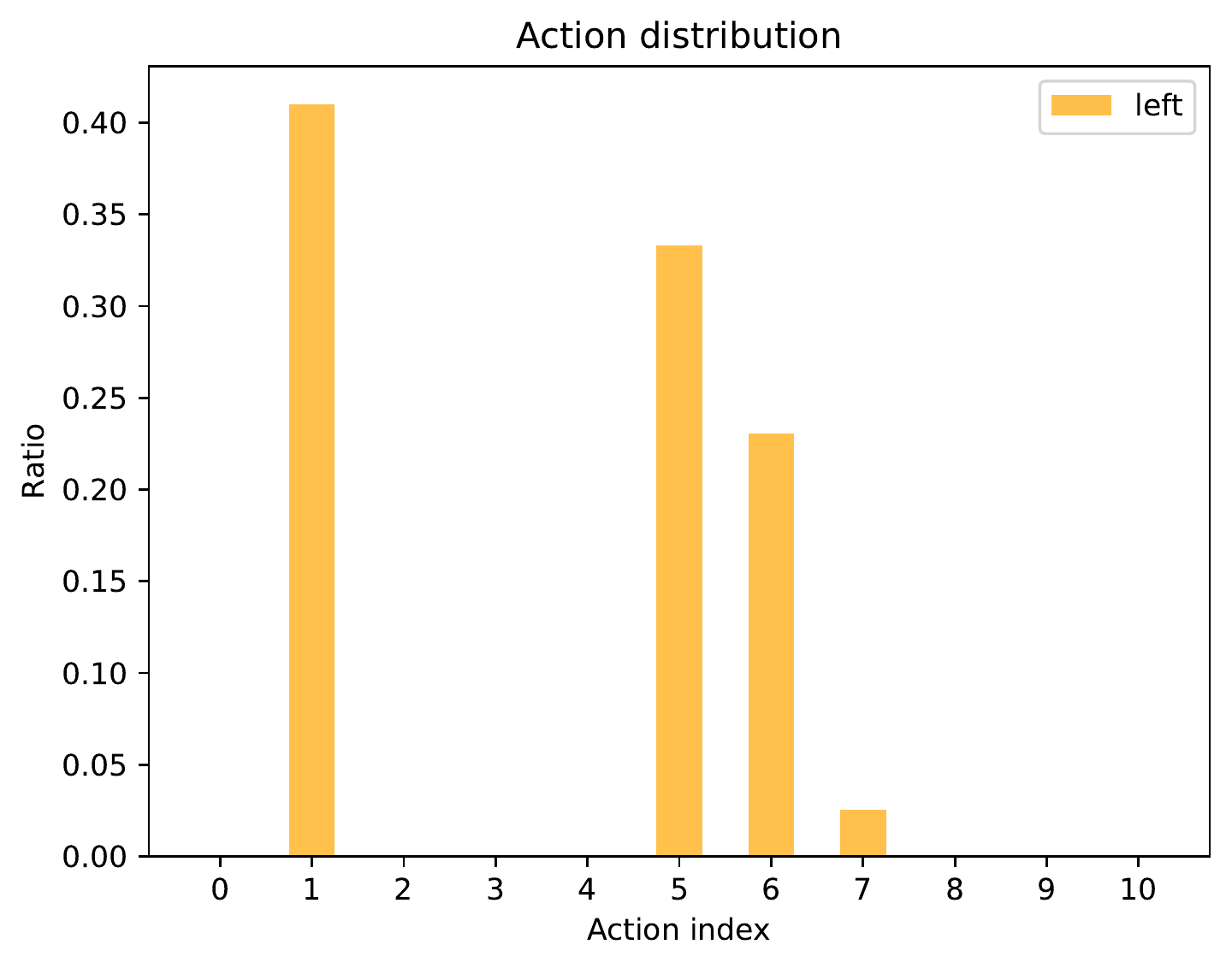} 
		\includegraphics[width=2.1in,height=0.7in]{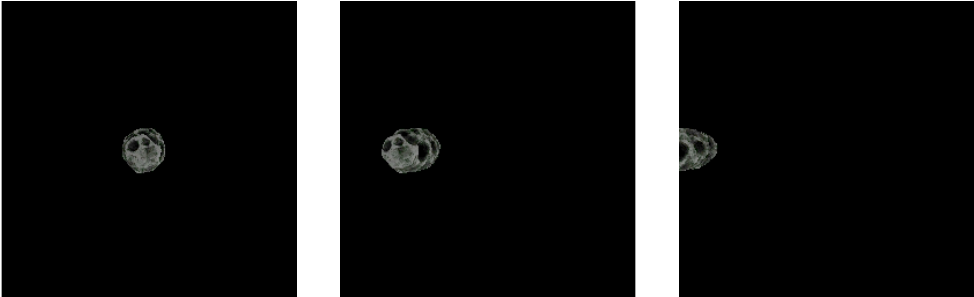} \label{fig16_3}}
	
		\subfloat[Asteroid06 with right translation]{\includegraphics[width=1in, height=0.7in]{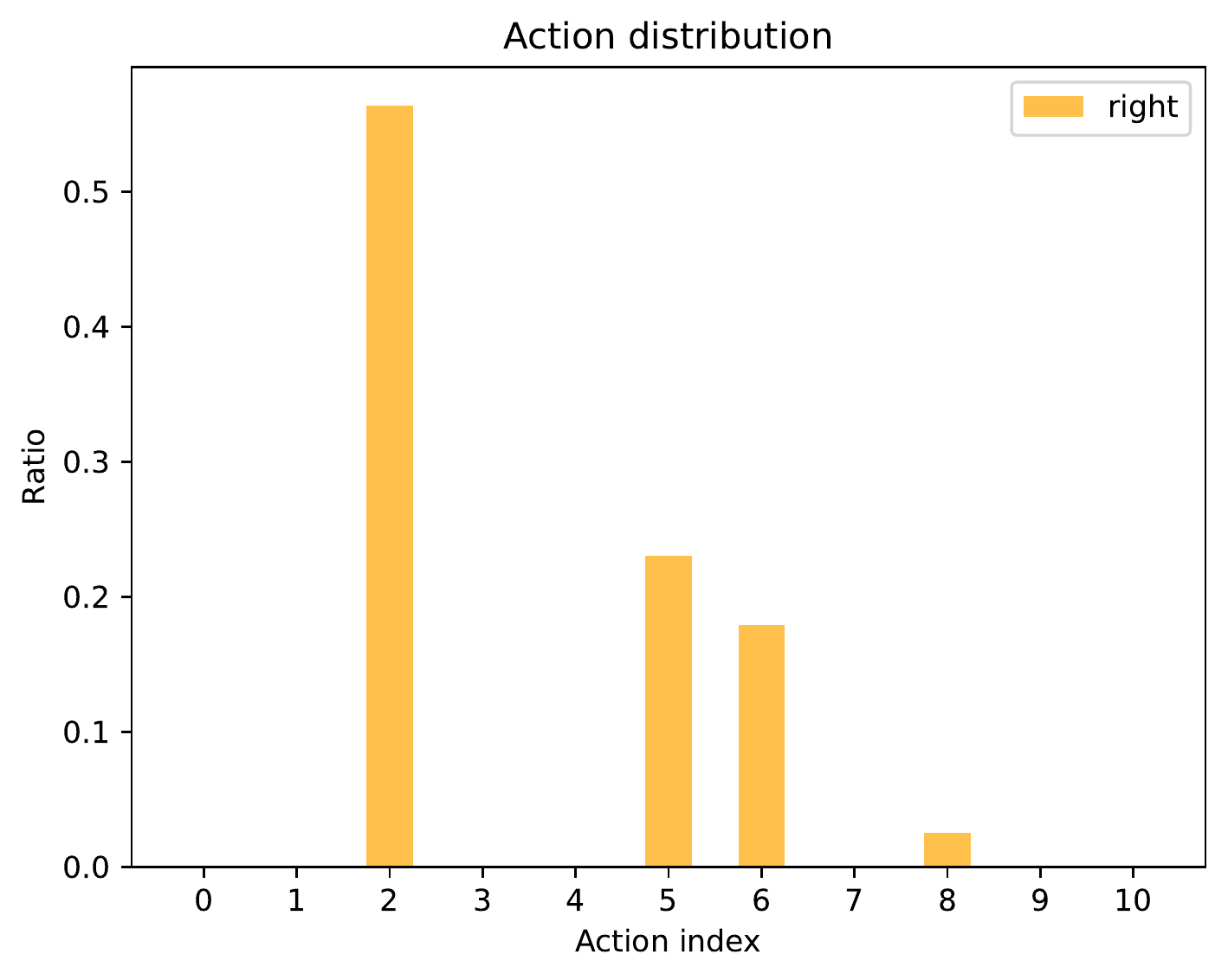} 
		\includegraphics[width=2.1in,height=0.7in]{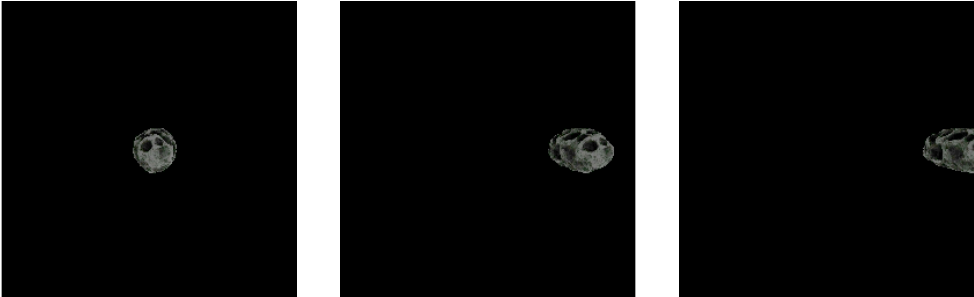}  \label{fig16_4}}
	\end{minipage}
	
	\begin{minipage}[b]{0.45\textwidth}
		\centering
		\subfloat[Up]{\includegraphics[width=0.5\textwidth]{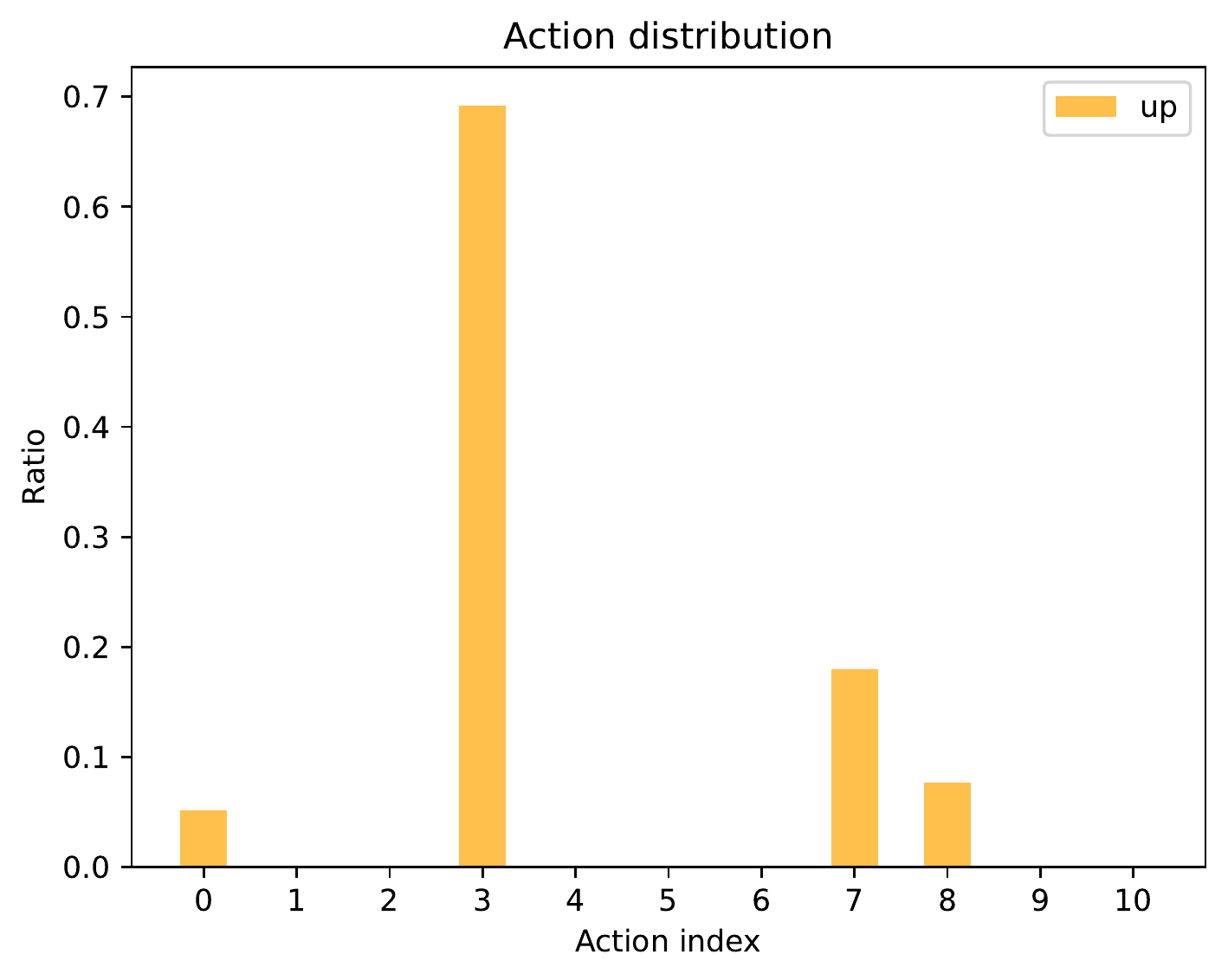}} \hfil
		\subfloat[Down]{\includegraphics[width=0.5\textwidth]{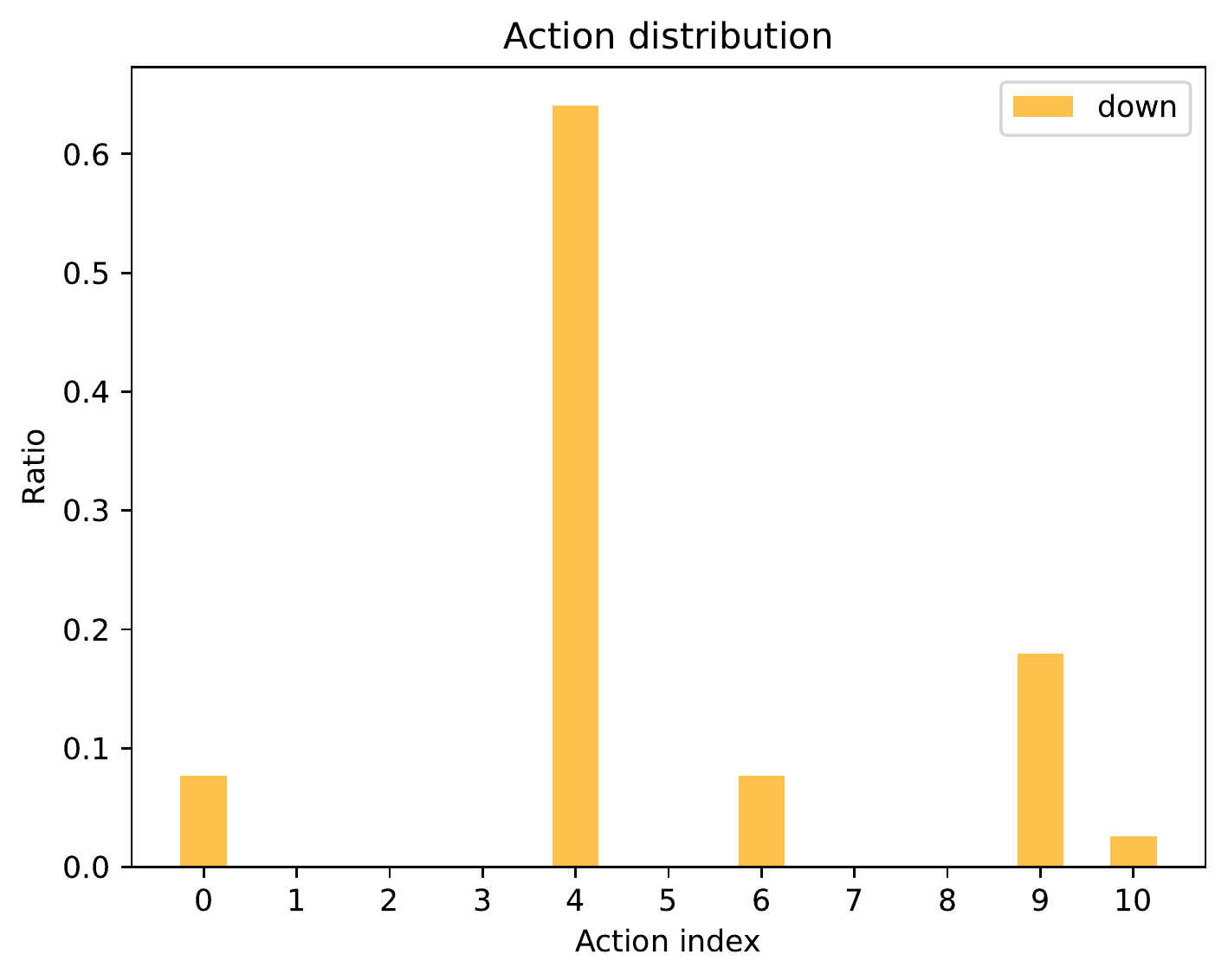}} \vfil
		\subfloat[Forward]{\includegraphics[width=0.5\textwidth]{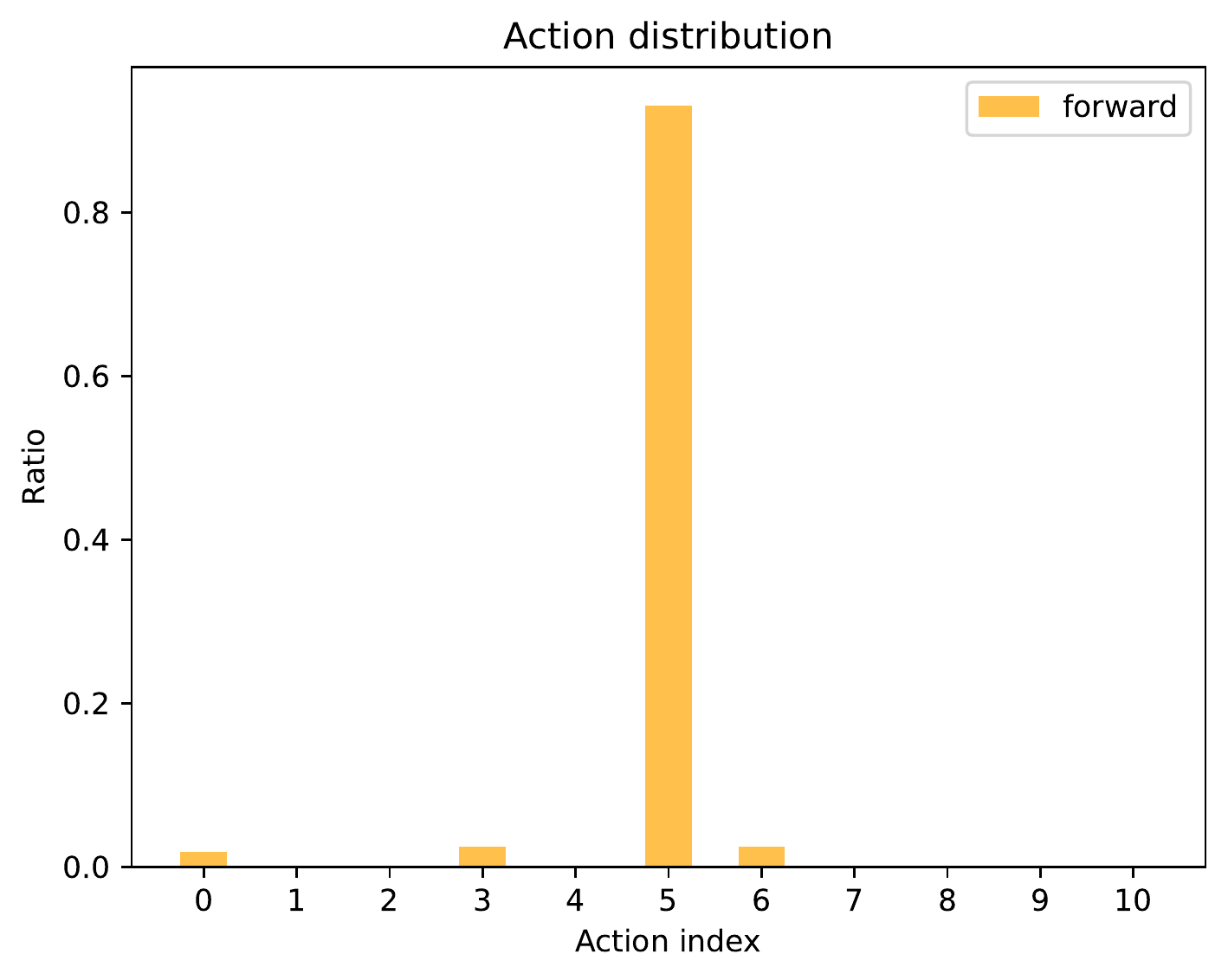}} \hfil
		\subfloat[Backward translation]{\includegraphics[width=0.5\textwidth]{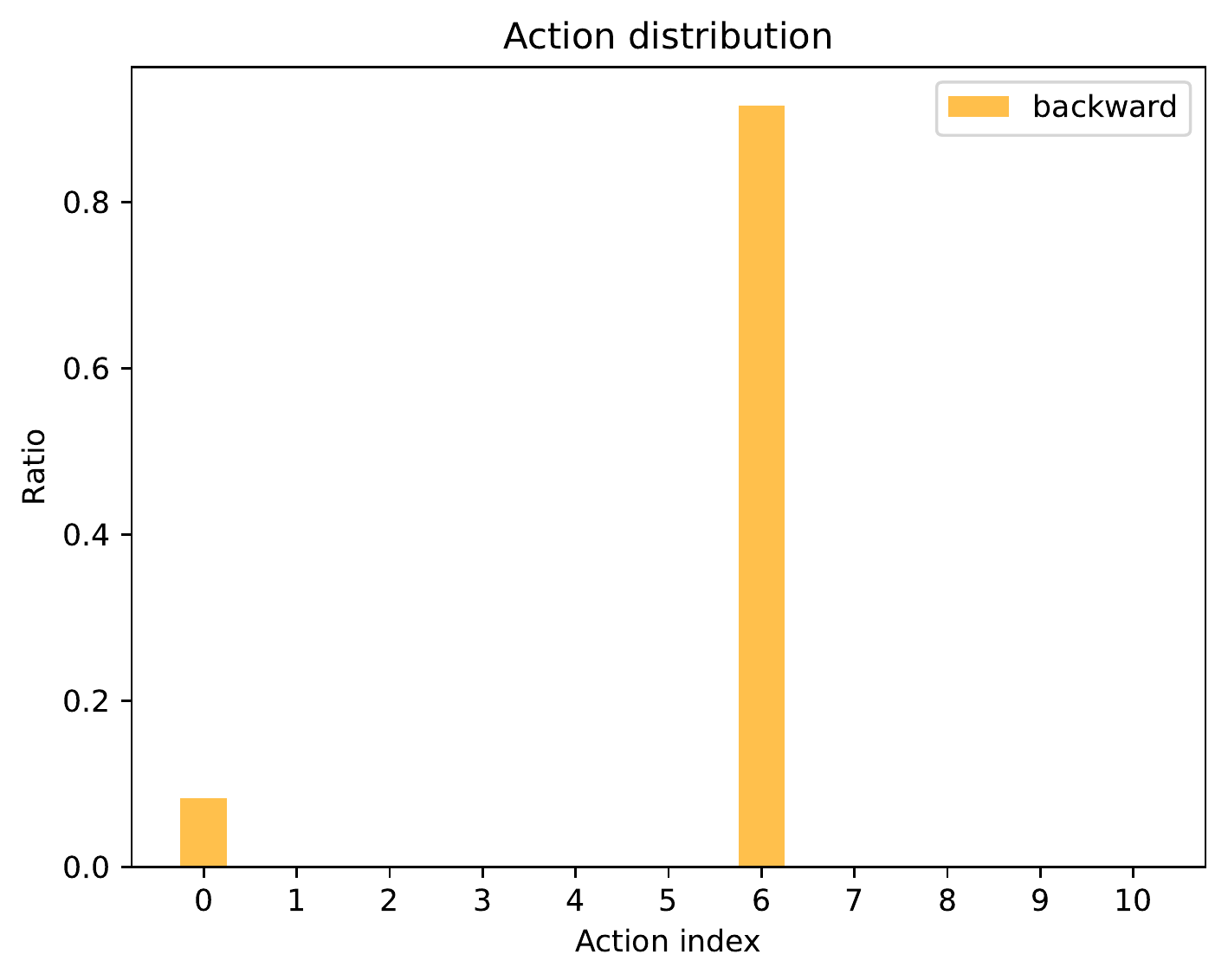}} 
	\end{minipage}
	\caption{The action statistics of DRLAVT to different motion patterns.}
	\label{fig16}
\end{figure}

\subsubsection{Exploration Inner Mechanism of DRLAVT}
Although we have proven the effectiveness and advancement of our method thereinbefore, the most interesting point is what agent actually learns from large-scale trial-and-errors with deep reinforcement learning. To this end, we initialize the non-cooperative target at the desired position (i.e. $X^{B}_{T} = \{0, 0, 5\}$) with respect to chaser and simplify its motion with 6 types of patterns, involving left, right, up, down, forward, and backward. Then, we keep chaser stationary and make statistics of actions selected by DRLAVT to each motion pattern which are plotted in Fig \ref{fig16}. It is obvious that our method take near-optimal actions to corresponding motion pattern. In addition, the geometry, texture and category of target just have negligible influences on DRLAVT, comparing Fig \ref{fig16_1}, \ref{fig16_2} with Fig \ref{fig16_3}, \ref{fig16_4}. Therefore, we think our method learnt the motion patterns of space non-cooperative target from images indeed.


\section{CONCLUSION \label{section6}}
In this paper, we construct an open-source simulated environment with 18 types of space non-cooperative object to facilitate the research of active visual tracking, especially for deep reinforcement learning based method. An end-to-end active visual tracker, DRLAVT, is proposed, which guides chasing spacecraft approach to arbitrary target by using color or RGBD images. Comparing to the PBVS baseline algorithm that adopts state-of-the-art SiamRPN tracker, our method achieves excellent performance even under severe disturbations. The impacts of network architectures, reward functions and multiple targets on DRLAVT are further studied, which provide significant reference for agent design based on deep reinforcement learning. At final, we prove that our DRLAVT indeed learnt the motion patterns of non-cooperative targets from extensive trial-and-errors. In future, we will evaluate our DRLAVT algorithm with ground physical simulation on planar gravity-offset testbed. Meanwhile, the intelligence of agent for active visual tracking task can be improved by competitive manner. 

\section*{APPENDIX}
\section{The perspective camera matrix}
In section \ref{section2}, we have mentioned that camera matrix is very important for active visual tracking. However, physics engine only provides perspective angle $(\alpha_x, \alpha_y)$ and resolution $(W, H)$. To this end, we derive the camera matrix by perspective projection principle, which is also illustrated in Fig. \ref{fig4}. One is 

Suppose that the virtual focus of perspective camera in both x and y axes are $(f_x, f_y)$, the size of image plane is $(w, h)$, and an object point $P=(X, Y, Z)$ in camera coordinate system is projected at $p=(x_i, y_i)$ in image plane. Fig. \ref{fig3} clearly shows that:
\begin{equation}
    \frac{X}{Z} = \frac{x_i}{f_x} \label{eq_29}
\end{equation}
and, 
\begin{equation}
    \frac{w/2}{f_x} = \tan\left( \frac{\alpha_x}{2}\right) \label{eq_30}
\end{equation}
Meanwhile, the transformation from image coordinate system to pixel coordinate system in x axis is formulated as:
\begin{align}
    u &= (x_i + \frac{w}{2}) \cdot \frac{W}{w} \label{eq_31} \\ 
    \nonumber &=x_i \cdot \frac{W}{w} + \frac{W}{2} 
\end{align}
Substitute Eq. \ref{eq_29} and \ref{eq_30} into Eq. \ref{eq_31}, it can be obtained:
\begin{equation}
    u = \frac{W}{ 2 \tan(\frac{\alpha_x}{2})Z} X+  \frac{W}{2} \label{eq_32}
\end{equation}
in which, the parameter $f_x$ is eliminated. Similarly, the transformation in y-axis direction from camera coordinate system to pixel coordinate system is also obtained:
\begin{equation}
    v = \frac{H}{ 2 \tan(\frac{\alpha_y}{2})Z} Y+  \frac{H}{2} \label{eq_33}
\end{equation}

We further rewrite Eq. \ref{eq_32} and \ref{eq_33} in homogeneous matrix form:
\begin{align}
    \begin{bmatrix}
       u \\
       v \\
       1
    \end{bmatrix}=
    \begin{bmatrix}
       \frac{W}{ 2 \tan(\frac{\alpha_x}{2})Z} & 0 & \frac{W}{2Z} \\
       0 &   \frac{H}{ 2 \tan(\frac{\alpha_y}{2})Z} & \frac{H}{2Z}\\
       0 & 0 & 1/Z
    \end{bmatrix} 
    \begin{bmatrix}
       X \\
       Y \\
       Z
    \end{bmatrix} \label{eq_34}
\end{align}
To eliminate the $Z$ variable in the transformation matrix, we multiply both sides of Eq. \ref{eq_34} by $Z$:
\begin{align}
    Z\begin{bmatrix}
       u \\
       v \\
       1
    \end{bmatrix}=
    \begin{bmatrix}
       \frac{W}{2 \tan(\alpha_x/2)} & 0 & \frac{W}{2} \\
       0 &   \frac{H}{ 2 \tan(\alpha_y/2)} & \frac{H}{2}\\
       0 & 0 & 1
    \end{bmatrix} 
    \begin{bmatrix}
       X \\
       Y \\
       Z
    \end{bmatrix} 
\end{align}

We therefore define camera intrinsic matrix $M_{intr}$ as following:
\begin{equation}
    M_{intr}=
    \begin{bmatrix}
       \frac{W}{2 \tan(\alpha_x/2)} & 0 & \frac{W}{2} \\
       0 &   \frac{H}{ 2 \tan(\alpha_y/2)} & \frac{H}{2}\\
       0 & 0 & 1
    \end{bmatrix} 
 \end{equation}


\bibliographystyle{IEEEtran}
\bibliography{IEEEabrv, Reinforcement_Learning, Object_Tracking, My_Work, Visuomotor_control}

\begin{thebibliography}{10}
\providecommand{\url}[1]{#1}
\csname url@samestyle\endcsname
\providecommand{\newblock}{\relax}
\providecommand{\bibinfo}[2]{#2}
\providecommand{\BIBentrySTDinterwordspacing}{\spaceskip=0pt\relax}
\providecommand{\BIBentryALTinterwordstretchfactor}{4}
\providecommand{\BIBentryALTinterwordspacing}{\spaceskip=\fontdimen2\font plus
\BIBentryALTinterwordstretchfactor\fontdimen3\font minus
  \fontdimen4\font\relax}
\providecommand{\BIBforeignlanguage}[2]{{%
\expandafter\ifx\csname l@#1\endcsname\relax
\typeout{** WARNING: IEEEtran.bst: No hyphenation pattern has been}%
\typeout{** loaded for the language `#1'. Using the pattern for}%
\typeout{** the default language instead.}%
\else
\language=\csname l@#1\endcsname
\fi
#2}}
\providecommand{\BIBdecl}{\relax}
\BIBdecl

\bibitem{huangDexterousTetheredSpace2017}
P.~Huang, F.~Zhang, J.~Cai, D.~Wang, Z.~Meng, and J.~Guo, ``Dexterous
  {Tethered} {Space} {Robot}: {Design}, {Measurement}, {Control}, and
  {Experiment},'' \emph{IEEE Transactions on Aerospace and Electronic Systems},
  vol.~53, no.~3, pp. 1452--1468, Jun. 2017.

\bibitem{flores-abadReviewSpaceRobotics2014}
A.~Flores-Abad, O.~Ma, K.~Pham, and S.~Ulrich, ``\BIBforeignlanguage{en}{A
  review of space robotics technologies for on-orbit servicing},''
  \emph{\BIBforeignlanguage{en}{Progress in Aerospace Sciences}}, vol.~68, pp.
  1--26, Jul. 2014.

\bibitem{dongAutonomousRoboticCapture2016}
{G. Dong} and Z.~H. Zhu, ``Autonomous robotic capture of non-cooperative target
  by adaptive extended kalman filter based visual servo,'' \emph{Acta
  Astronautica}, vol. 122, pp. 209--218, 2016.

\bibitem{petitVisionbasedDetectionTracking2012}
A.~Petit, E.~Marchand, and K.~Kanani, ``Vision-based detection and tracking for
  space navigation in a rendezvous context,'' in \emph{Int. Symp. on Artificial
  Intelligence, Robotics and Automation in Space, i-SAIRAS}, 2012.

\bibitem{ramosVisionbasedTrackingNoncooperative2018}
J.~H. Ramos, T.~D. Woodbury, and J.~E. Hurtado, ``Vision-based tracking of
  non-cooperative space bodies to support active attitude control detection,''
  in \emph{2018 {AIAA} {SPACE} and {Astronautics} {Forum} and
  {Exposition}}.\hskip 1em plus 0.5em minus 0.4em\relax American Institute of
  Aeronautics and Astronautics, 2018.

\bibitem{zhou3DVisualTracking2021}
D.~Zhou, G.~Sun, and X.~Hong, ``{3D} {Visual} {Tracking} {Framework} with
  {Deep} {Learning} for {Asteroid} {Exploration},'' \emph{arXiv:2111.10737
  [cs]}, Nov. 2021, arXiv: 2111.10737.

\bibitem{fourieFlightResultsVisionBased2014}
D.~Fourie, B.~E. Tweddle, S.~Ulrich, and A.~Saenz-Otero, ``Flight {Results} of
  {Vision}-{Based} {Navigation} for {Autonomous} {Spacecraft} {Inspection} of
  {Unknown} {Objects},'' \emph{Journal of Spacecraft and Rockets}, vol.~51,
  no.~6, pp. 2016--2026, 2014.

\bibitem{zhangImprovedRealtimeVisual2016}
L.~Zhang, F.~Zhu, and Y.~Hao, ``An improved real-time visual tracking method
  for space non-cooperative target,'' in \emph{Infrared Technology and
  Applications, and Robot Sensing and Advanced Control}, vol. 10157, 2016, p.
  1015733.

\bibitem{volpePassiveCameraBased2018}
R.~Volpe, G.~B. Palmerini, and M.~Sabatini, ``\BIBforeignlanguage{en}{A passive
  camera based determination of a non-cooperative and unknown satellite's pose
  and shape},'' \emph{\BIBforeignlanguage{en}{Acta Astronautica}}, vol. 151,
  pp. 805--817, Oct. 2018.

\bibitem{zhou2DVisionbasedTracking2021}
D.~Zhou, G.~Sun, J.~Song, and W.~Yao, ``\BIBforeignlanguage{en}{{2D}
  vision-based tracking algorithm for general space non-cooperative objects},''
  \emph{\BIBforeignlanguage{en}{Acta Astronautica}}, vol. 188, pp. 193--202,
  Nov. 2021.

\bibitem{dongPositionbasedVisualServo2015a}
G.~Dong and Z.~Zhu, ``Position-based visual servo control of autonomous robotic
  manipulators,'' \emph{Acta Astronautica}, vol. 115, pp. 291--302, 2015.

\bibitem{sunAdaptiveRelativePose2018}
L.~Sun and Z.~Zheng, ``Adaptive relative pose control of spacecraft with model
  couplings and uncertainties,'' \emph{Acta Astronautica}, vol. 143, pp.
  29--36, 2018.

\bibitem{liuRobustAdaptiveRelative2020}
J.~Liu, H.~Li, Y.~Luo, and J.~Zhang, ``Robust adaptive relative position and
  attitude integrated control for approaching uncontrolled tumbling
  spacecraft,'' \emph{Proceedings of the Institution of Mechanical Engineers,
  Part G: Journal of Aerospace Engineering}, vol. 234, no.~2, pp. 361--374,
  2020.

\bibitem{wangEyeinHandTrackingControl2017}
H.~Wang, D.~Guo, H.~Xu, W.~Chen, T.~Liu, and K.~K. Leang, ``Eye-in-{{Hand
  Tracking Control}} of a {{Free}}-{{Floating Space Manipulator}},'' \emph{IEEE
  Transactions on Aerospace and Electronic Systems}, vol.~53, no.~4, pp.
  1855--1865, 2017.

\bibitem{felicettiImagebasedAttitudeManeuvers2018}
L.~Felicetti and M.~R. Emami, ``\BIBforeignlanguage{en}{Image-based attitude
  maneuvers for space debris tracking},''
  \emph{\BIBforeignlanguage{en}{Aerospace Science and Technology}}, vol.~76,
  pp. 58--71, May 2018.

\bibitem{zhaoImagebasedControlRendezvous2021}
X.~Zhao, M.~R. Emami, and S.~Zhang, ``\BIBforeignlanguage{en}{Image-based
  control for rendezvous and synchronization with a tumbling space debris},''
  \emph{\BIBforeignlanguage{en}{Acta Astronautica}}, vol. 179, pp. 56--68, Feb.
  2021.

\bibitem{henriquesHighSpeedTrackingKernelized2015}
J.~F. Henriques, R.~Caseiro, P.~Martins, and J.~Batista, ``High-{Speed}
  {Tracking} with {Kernelized} {Correlation} {Filters},'' \emph{IEEE
  Transactions on Pattern Analysis and Machine Intelligence}, vol.~37, no.~3,
  pp. 583--596, Mar. 2015.

\bibitem{liHighPerformanceVisual2018}
B.~Li, J.~Yan, W.~Wu, Z.~Zhu, and X.~Hu, ``High performance visual tracking
  with siamese region proposal network,'' in \emph{Proceedings of the IEEE
  conference on computer vision and pattern recognition}, 2018, pp. 8971--8980.

\bibitem{levineLearningContactrichManipulation2015}
S.~Levine, N.~Wagener, and P.~Abbeel, ``Learning contact-rich manipulation
  skills with guided policy search,'' in \emph{2015 {{IEEE International
  Conference}} on {{Robotics}} and {{Automation}} ({{ICRA}})}, 2015, pp.
  156--163.

\bibitem{levineEndtoendTrainingDeep2016}
S.~Levine, C.~Finn, T.~Darrell, and P.~Abbeel, ``End-to-end training of deep
  visuomotor policies,'' \emph{The Journal of Machine Learning Research},
  vol.~17, no.~1, pp. 1334--1373, Jan. 2016.

\bibitem{loquercioDeepDroneRacing2020}
A.~Loquercio, E.~Kaufmann, R.~Ranftl, A.~Dosovitskiy, V.~Koltun, and
  D.~Scaramuzza, ``Deep {Drone} {Racing}: {From} {Simulation} to {Reality}
  {With} {Domain} {Randomization},'' \emph{IEEE Transactions on Robotics},
  vol.~36, no.~1, pp. 1--14, Feb. 2020.

\bibitem{suttonReinforcementLearningIntroduction2018}
R.~S. Sutton and A.~G. Barto, \emph{Reinforcement learning: An
  introduction}.\hskip 1em plus 0.5em minus 0.4em\relax MIT press, 2018.

\bibitem{briangaudetAdaptivePinpointFuel2014}
G.~Brian and F.~Roberto, ``Adaptive pinpoint and fuel efficient mars landing
  using reinforcement learning,'' \emph{IEEE/CAA Journal of Automatica Sinica},
  vol.~1, no.~4, pp. 397--411, Oct. 2014.

\bibitem{scorsoglioImagebasedDeepReinforcement}
A.~Scorsoglio, R.~Furfaro, R.~Linares, and B.~Gaudet, ``Image-based deep
  reinforcement learning for autonomous lunar landing,'' in \emph{AIAA Scitech
  2020 Forum}, 2020, p. 1910.

\bibitem{sanchez-sanchezRealTimeOptimalControl2018}
C.~Sánchez-Sánchez and D.~Izzo, ``Real-{Time} {Optimal} {Control} via {Deep}
  {Neural} {Networks}: {Study} on {Landing} {Problems},'' \emph{Journal of
  Guidance, Control, and Dynamics}, vol.~41, no.~5, pp. 1122--1135, 2018.

\bibitem{mnihHumanlevelControlDeep2015}
V.~Mnih, K.~Kavukcuoglu, D.~Silver, and et~al.,
  ``\BIBforeignlanguage{en}{Human-level control through deep reinforcement
  learning},'' \emph{\BIBforeignlanguage{en}{Nature}}, vol. 518, no. 7540, pp.
  529--533, Feb. 2015.

\bibitem{harris2019spacecraft}
A.~Harris, T.~Teil, and H.~Schaub, ``Spacecraft decision-making autonomy using
  deep reinforcement learning,'' in \emph{29th {AAS}/{AIAA} space flight
  mechanics meeting, hawaii}, 2019, pp. 1--19.

\bibitem{hovellDeepReinforcementLearning2020}
{K. Hovell} and S.~Ulrich, ``On {Deep} {Reinforcement} {Learning} for
  {Spacecraft} {Guidance},'' in \emph{{AIAA} {Scitech} 2020 {Forum}}.\hskip 1em
  plus 0.5em minus 0.4em\relax American Institute of Aeronautics and
  Astronautics, 2020.

\bibitem{kirkhovellDeepReinforcementLearning2021}
K.~Hovell and S.~Ulrich, ``\BIBforeignlanguage{en}{Deep {Reinforcement}
  {Learning} for {Spacecraft} {Proximity} {Operations} {Guidance}},''
  \emph{\BIBforeignlanguage{en}{Journal of Spacecraft and Rockets}}, Jan. 2021.

\bibitem{barth-maronDistributedDistributionalDeterministic2018}
G.~Barth-Maron, M.~W. Hoffman, D.~Budden, W.~Dabney, D.~Horgan, D.~TB,
  A.~Muldal, N.~Heess, and T.~Lillicrap, ``Distributed {Distributional}
  {Deterministic} {Policy} {Gradients},'' \emph{arXiv:1804.08617 [cs, stat]},
  Apr. 2018, arXiv: 1804.08617.

\bibitem{bousmalisUsingSimulationDomain2018}
K.~Bousmalis, A.~Irpan, P.~Wohlhart, Y.~Bai, M.~Kelcey, M.~Kalakrishnan,
  L.~Downs, J.~Ibarz, P.~Pastor, K.~Konolige, S.~Levine, and V.~Vanhoucke,
  ``Using {Simulation} and {Domain} {Adaptation} to {Improve} {Efficiency} of
  {Deep} {Robotic} {Grasping},'' in \emph{2018 {IEEE} {International}
  {Conference} on {Robotics} and {Automation} ({ICRA})}, May 2018, pp.
  4243--4250, iSSN: 2577-087X.

\bibitem{sunderhaufLimitsPotentialsDeep2018}
N.~Sünderhauf, O.~Brock, W.~Scheirer, R.~Hadsell, D.~Fox, J.~Leitner,
  B.~Upcroft, P.~Abbeel, W.~Burgard, M.~Milford, and P.~Corke,
  ``\BIBforeignlanguage{en}{The limits and potentials of deep learning for
  robotics},'' \emph{\BIBforeignlanguage{en}{The International Journal of
  Robotics Research}}, vol.~37, no. 4-5, pp. 405--420, Apr. 2018.

\bibitem{luoEndtoEndActiveObject2020}
W.~Luo, P.~Sun, F.~Zhong, W.~Liu, T.~Zhang, and Y.~Wang, ``End-to-{End}
  {Active} {Object} {Tracking} and {Its} {Real}-{World} {Deployment} via
  {Reinforcement} {Learning},'' \emph{IEEE Transactions on Pattern Analysis and
  Machine Intelligence}, vol.~42, no.~6, pp. 1317--1332, Jun. 2020.

\bibitem{huangGOT10kLargeHighDiversity2019}
L.~Huang, X.~Zhao, and K.~Huang, ``{GOT}-10k: {A} {Large} {High}-{Diversity}
  {Benchmark} for {Generic} {Object} {Tracking} in the {Wild},'' \emph{IEEE
  Transactions on Pattern Analysis and Machine Intelligence}, pp. 1--1, 2019.

\bibitem{wuObjectTrackingBenchmark2015}
Y.~Wu, J.~Lim, and M.-H. Yang, ``Object {Tracking} {Benchmark},'' \emph{IEEE
  Transactions on Pattern Analysis and Machine Intelligence}, vol.~37, no.~9,
  pp. 1834--1848, Sep. 2015.

\bibitem{kristanSeventhVisualObject2019}
M.~Kristan, J.~Matas, A.~Leonardis, M.~Felsberg, and et~al., ``The {Seventh}
  {Visual} {Object} {Tracking} {VOT2019} {Challenge} {Results},'' \emph{Proc.
  IEEE/CVF Int. Conf. Comput. Vis. Work.}, 2019.

\bibitem{krizhevsky2017imagenet}
A.~Krizhevsky, I.~Sutskever, and G.~E. Hinton, ``Imagenet classification with
  deep convolutional neural networks,'' \emph{Communications of the ACM},
  vol.~60, no.~6, pp. 84--90, 2017.

\bibitem{he2016deep}
K.~He, X.~Zhang, S.~Ren, and J.~Sun, ``Deep residual learning for image
  recognition,'' in \emph{Proceedings of the IEEE conference on computer vision
  and pattern recognition}, 2016, pp. 770--778.

\end{thebibliography}

\newpage

\begin{IEEEbiography}[{\includegraphics[width=0.9in,height=1.25in,clip]{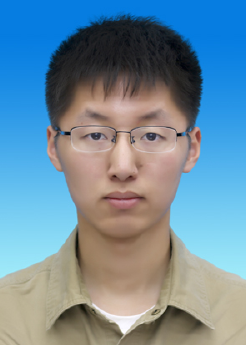}}]{Dong Zhou}
	was born in Hunan, China, in 1996. He received the B.S degree in automation from Harbin Engineering University, Harbin, China, in 2018. He is currently working toward the Ph.D. degree in the Department of Control Science and Engineering, Harbin Institute of Technology, Harbin, China. His research interests include space non-cooperative object visual tracking, 3D computer vision, and deep reinforcement learning. 
\end{IEEEbiography}


\begin{IEEEbiography}[{\includegraphics[width=0.9in,height=1.25in,clip,keepaspectratio]{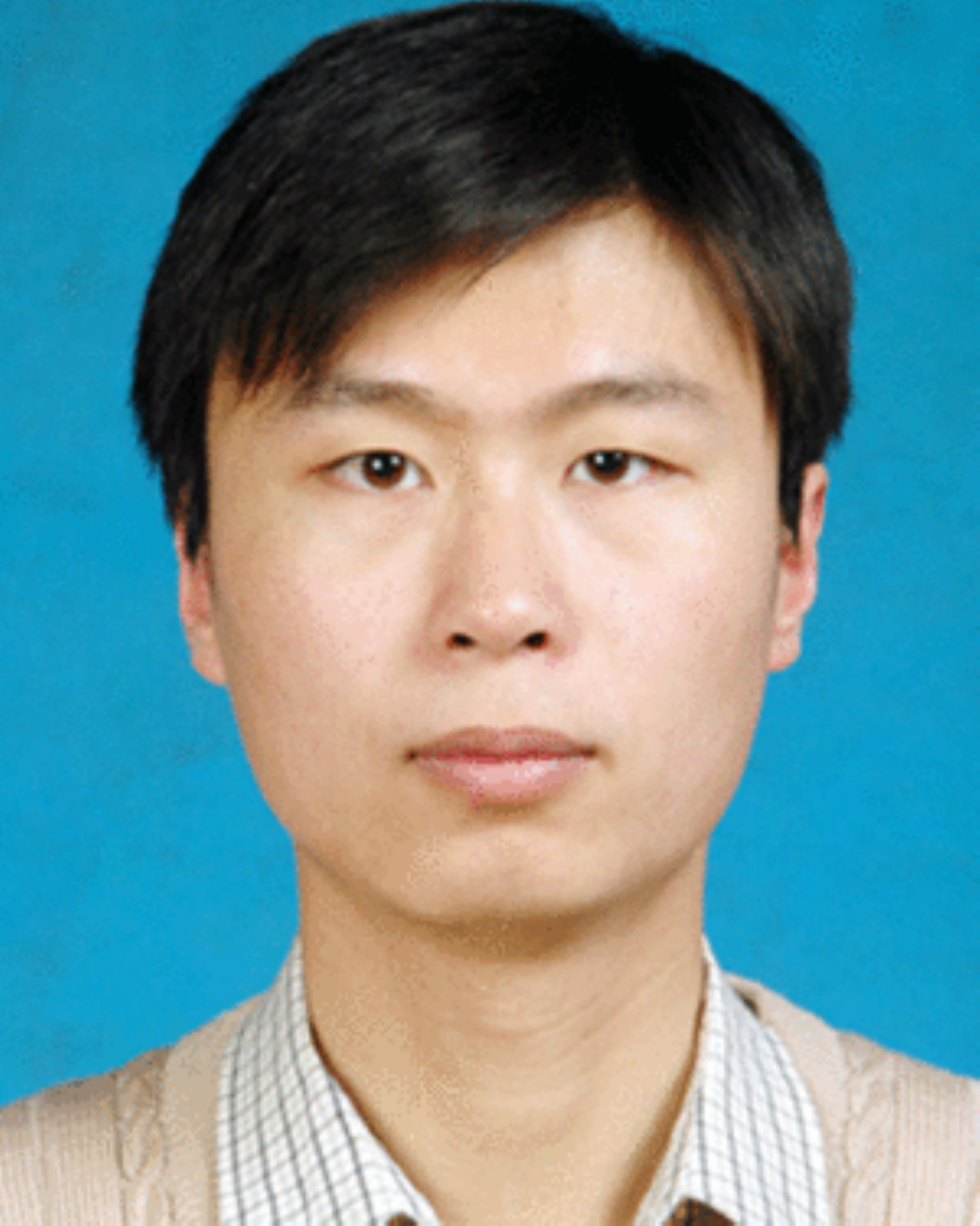}}]{Guanghui Sun}
	 was born in Henan Province, China, in 1983. He received the B.S. degree in Automation from Harbin Institute of Technology, Harbin, China, in 2005, and the M.S. and Ph.D. degrees in Control Science and Engineering from Harbin Institute of Technology, Harbin, China, in 2007 and 2010, respectively. He is currently a professor with Department of Control Science and Engineering in Harbin Institute of Technology, Harbin, China. His research interests include machine learning, computer vision, and aerospace technology.
\end{IEEEbiography}

\begin{IEEEbiography}[{\includegraphics[width=0.9in,height=1.25in,clip,keepaspectratio]{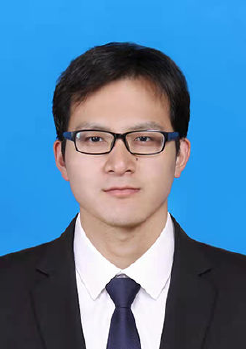}}]{Wenxiao Lei}
    received the B.S. degree in automation and M.S. degree in control science and engineering from Harbin Institute of Technology, Harbin, China, in 2016 and 2019, repectively. He is currently working toward the Ph.D. degree in the Department of Control Science and Engineering, Harbin Institute of Technology, Harbin, China. His research interests include intelligent control of robotic manipulators, robotic motion planning, non-cooperative target capture.
\end{IEEEbiography}

\end{document}